\documentclass[11pt, a4paper, logo, twocolumn]{craftjarvis}

\ifx\pdfinfoomitdate\undefined\else
  \pdftrailerid{redacted}
\fi

\makeatletter
\renewcommand\bibentry[1]{\nocite{#1} {\frenchspacing\@nameuse{BR@r@#1\@extra@b@citeb}}}
\makeatother

\usepackage{xcolor}
\definecolor{Gray}{gray}{0.9}
\definecolor{mygreen}{rgb}{0.0, 0.5, 0.0}
\definecolor{myred}{rgb}{0.8, 0.25, 0.33}
\definecolor{myblue}{rgb}{0.19, 0.55, 0.91}
\definecolor{uclablue}{rgb}{0.15, 0.45, 0.68}
\definecolor{boxgreen}{rgb}{0.02, 0.66, 0.02}
\definecolor{boxred}{rgb}{0.66, 0.1, 0.1}
\definecolor{boxblue}{rgb}{0.01, 0.01, 0.73}

\definecolor{mygray}{gray}{0.4}

\usepackage{times}
\usepackage{graphicx}
\usepackage{amssymb}
\usepackage{url}
\usepackage{microtype}
\usepackage{booktabs}
\usepackage{amsmath}
\usepackage{mathtools}
\usepackage{amsthm}
\usepackage{algorithm}
\usepackage{algpseudocode}
\usepackage{pifont}
\usepackage{wrapfig}
\usepackage{colortbl}
\usepackage{multirow}
\usepackage{makecell}
\usepackage{enumitem}
\usepackage{tabularx}
\usepackage{array}
\usepackage{amsfonts}
\usepackage{nicefrac}
\usepackage{xspace}
\usepackage{mdframed}
\usepackage[export]{adjustbox}

\newcolumntype{Y}{>{\arraybackslash}X}

\renewcommand{\paragraph}[1]{\noindent\textbf{#1.}}

\makeatletter
\DeclareRobustCommand\onedot{\futurelet\@let@token\@onedot}
\def\@onedot{\ifx\@let@token.\else.\null\fi\xspace}

\makeatother

\usepackage[authoryear, sort&compress, round]{natbib}

\usepackage{hyperref}
\usepackage{marvosym}
\usepackage[capitalize,noabbrev]{cleveref}

\newcommand{\alg}{PGT\xspace}

\theoremstyle{plain}
\newtheorem{theorem}{Theorem}[section]
\newtheorem{proposition}[theorem]{Proposition}

\theoremstyle{definition}

\theoremstyle{remark}

\title{Preference Goal Tuning: Post-Training as Latent Control for Frozen Policies}

\reportnumber{}
\paperurl={https://arxiv.org/abs/2412.02125}

\author[1$\dag$]{Guangyu~Zhao}
\author[2$\dag$]{Kewei~Lian}
\author[1$\dag$]{Haoxuan~Ru}
\author[1$\dag$]{Borong~Zhang}
\author[1]{Haowei~Lin}
\author[1]{Zhancun~Mu}
\author[3]{Haobo~Fu}
\author[3]{Qiang~Fu}
\author[1]{Shaofei~Cai}
\author[1]{Zihao~Wang}
\author[1$\textrm{\Letter}$]{Yitao~Liang}

\affil[1]{Peking University}
\affil[2]{National University of Singapore}
\affil[3]{Tencent AI Lab}

\begin{abstract}
Goal-conditioned policies enable decision-making models to execute diverse behaviors based on specified goals, yet their downstream performance is often highly sensitive to the choice of instructions or prompts. To bypass the limitations of discrete text prompts, we formulate post-training adaptation as a latent control problem, where the goal embedding serves as a continuous control variable to modulate the behavior of a frozen policy. We propose Preference Goal Tuning (PGT), a framework that optimizes this latent control variable to align the induced trajectory distribution with task preferences. Unlike standard fine-tuning that updates policy parameters, PGT keeps the policy frozen and updates only the latent goal using a trajectory-level preference objective. This approach essentially searches for the optimal conditioning input that maximizes the likelihood of preferred behaviors while suppressing undesirable ones. We evaluate PGT on the Minecraft SkillForge benchmark across 17 tasks. With minimal data, PGT achieves average relative improvements of 72.0\% and 81.6\% on two foundation policies, consistently outperforming expert-crafted prompts. Crucially, by decoupling task alignment (latent goal) from physical dynamics (frozen policy), PGT surpasses full fine-tuning by 13.4\% in out-of-distribution settings, demonstrating superior robustness and generalization.
\end{abstract}

\begin{document}

\correspondingauthor{Yitao~Liang, $\dag$ indicates co-first author \\
Guangyu~Zhao~<zhaogy24@stu.pku.edu.cn>,
Kewei~Lian~<liankewei@u.nus.edu>,
Yitao~Liang~<yitaol@pku.edu.cn>
}

\maketitle

\section{Introduction}

Goal-conditioned policies pretrained on large-scale datasets have demonstrated strong capabilities in interpreting instructions and executing corresponding behaviors~\citep{rt2, openvla, rocket1, humantorobot}. Such instructions, often referred to as ``prompts'', may take diverse forms, including text~\citep{steve1, openvla, gametars}, images or videos~\citep{mimicplay, groot}, and multimodal inputs~\citep{groot2, pappas2025navigation}.

Despite their generality, the downstream performance of instruction-following policies is often highly sensitive to the choice of prompts~\citep{steve1, jarvis1, openvla, deps}. Identifying effective prompts typically relies on manual trial and error in a discrete text space, and prompts that appear reasonable to humans do not necessarily elicit optimal behavior. 

Ideally, adaptation should effectively align the policy with specific tasks while preserving the robust, general-purpose capabilities acquired during pretraining. However, conventional approaches struggle to balance these needs. Prompt engineering respects the frozen policy but is limited by the discrete, derivative-free nature of text search, often failing to elicit optimal behaviors. Conversely, fine-tuning the policy with reinforcement learning or imitation learning can effectively shape behavior but risks catastrophic forgetting or overfitting to narrow task distributions, thereby shattering the broad generalization capabilities of the foundation model~\citep{rlgen, ptgm}. This dilemma highlights the need for an adaptation mechanism that offers the optimization power of fine-tuning while maintaining the structural stability of a frozen policy.

To bridge this gap, we adopt a \emph{latent control} perspective. We observe that a pretrained goal-conditioned policy defines a vast family of potential behaviors, indexed by its continuous goal embedding space. Rather than modifying the policy's weights—which encodes the agent's fundamental understanding of physics and skills—we propose to adapt behavior solely by optimizing the \emph{latent goal} that conditions the frozen policy. This formulation serves as a continuous, differentiable control interface, allowing for precise behavior modulation beyond what is possible with discrete text prompts. Furthermore, because the optimization is restricted to a low-dimensional latent vector rather than millions of policy parameters, this approach is inherently sample-efficient and resistant to overfitting.

Based on this formulation, we introduce \emph{\textbf{P}reference \textbf{G}oal-\textbf{T}uning} (\alg), a post-training framework that optimizes this latent control variable using preference learning. Unlike standard fine-tuning, \alg keeps the policy parameters frozen and treats the latent goal as a learnable parameter to reweight the induced trajectory distribution. Starting from an initial prompt, \alg collects a small number of trajectories, constructs pairwise preferences based on relative quality, and applies a contrastive objective to update the latent goal. This process effectively searches for the optimal conditioning input that maximizes the likelihood of preferred behaviors while suppressing undesirable ones.

We evaluate \alg on the \emph{Minecraft SkillForge} benchmark~\citep{malmo}, which provides a suitable testbed for task-level adaptation: tasks such as resource collection are semantically consistent yet executed across diverse environments. Across 17 tasks and two pretrained foundation policies, \alg consistently improves performance in both in-distribution and out-of-distribution settings, surpasses the best human-selected prompts, and demonstrates robustness to environmental variation. Finally, we explore \alg as an efficient approach to continual learning, where storing a compact latent per task enables scalable adaptation without catastrophic forgetting or task interference. We also demonstrate that \alg is a general post-training mechanism applicable to foundation policies across domains, not specific to Minecraft.

\section{Related Work}

\subsection{Goal-Conditioned Imitation Learning}

Goal-conditioned imitation learning (GCIL) extends imitation learning by conditioning policies on explicit goals, enabling a single policy to execute diverse tasks given different goal inputs.
Most prior GCIL methods are formulated under an end-to-end training paradigm, where both the policy and the goal representation are jointly optimized from expert demonstrations under a behavioral cloning objective.

One line of work represents goals as tokens or embeddings that are jointly modeled with states and actions in a sequence, as in large-scale instruction- or task-conditioned policies~\citep{bc-z, octo, openvla, pi0, pi05, rocket1, rocket2, dexgraspvla, gametars}.
Another class of approaches adopts explicit goal encoder--policy decoder architectures, where a learned goal representation conditions the policy network~\citep{mimicplay, steve1, rdt}.
A further line of work models goals as latent variables using variational objectives, such as conditional VAEs, jointly learning goal representations and policies in a self-supervised manner~\citep{groot, groot2}. These works demonstrate that goal-conditioned policies can support a wide range of tasks within a single model.
However, a common assumption underlying these methods is the joint or end-to-end optimization of both the policy and the goal representation. This assumption can make post-training adaptation costly and may introduce interference or degradation of generalization when adapting to new tasks or environments.

In contrast, our work departs from the standard GCIL paradigm.
We consider post-training adaptation of a pretrained goal-conditioned policy under a frozen policy backbone.
Rather than treating the goal representation as a passive conditioning input learned during pretraining, we explicitly optimize the latent goal embedding after pretraining using trajectory-level preference feedback.
This setting enables task-specific improvement while mitigating interference and preserving out-of-distribution generalization.

\subsection{Preference Learning}

Preference learning~\citep{slic, slichf, dpo, ipo, simpo} studies how to train models from comparative or ranked feedback.
It has become a central paradigm in reinforcement learning from human feedback (RLHF)~\citep{rlhf, lmrlhf}, where preference data is used to align model behavior, especially for large language models.
To avoid the need for training an explicit reward model, prior work has proposed preference-based learning methods that directly optimize model behavior from pairwise comparisons.
Direct Preference Optimization (DPO)~\citep{dpo} and its variants optimize models directly from preference pairs under a KL-regularized objective, enabling stable and efficient alignment with a reference model.
Several extensions, such as SLiC-HF~\citep{slic, slichf} and IPO~\citep{ipo}, further refine the preference objective to improve adherence to the reference policy. Beyond human-labeled preference pairs for aligning language models, preference learning has also been explored in sequential decision-making~\citep{apo} and reward-based pseudo pairs~\citep{grape}, where feedback is provided over a trajectory.

Most existing methods focus on the objective. In contrast, our work considers trajectory-level preference learning in goal-conditioned sequential decision-making and applies preference optimization to a different object: the latent goal embedding.
By adapting preference-based objectives to optimize only the goal representation under a frozen policy backbone, our approach enables efficient post-training improvement while preserving the behavior and generalization properties of the pretrained policy.

\subsection{Minecraft Agents and Open-World Evaluation}

Minecraft is a common testbed for studying large-scale embodied agents for its open-ended environment, task diversity, and support for long-horizon decision-making.
Prior work has developed agents and learning systems in Minecraft, including VPT~\citep{vpt}, STEVE-1~\citep{steve1}, GROOT~\citep{groot, groot2} and ROCKET~\cite{rocket1, rocket2, rocket3}, demonstrating capabilities such as skill acquisition and generalization.

These environments have been used to evaluate a range of challenges, including long-horizon control~\citep{jin2023mini, deps}, precise interaction~\citep{zhang2020high, vpt, rocket1}, and out-of-distribution generalization~\citep{yang2023essential, opencontrol}.
In this work, we adopt Minecraft as an evaluation platform to assess post-training adaptation of goal-conditioned policies across diverse tasks and environments.
We emphasize that our focus is not on architectural innovations for open-world exploration, but on evaluating whether optimizing latent goals can improve task performance and generalization without modifying the underlying policy.

\section{Methodology}

\begin{figure*}[htbp]
    \centering
    \includegraphics[width=\linewidth]{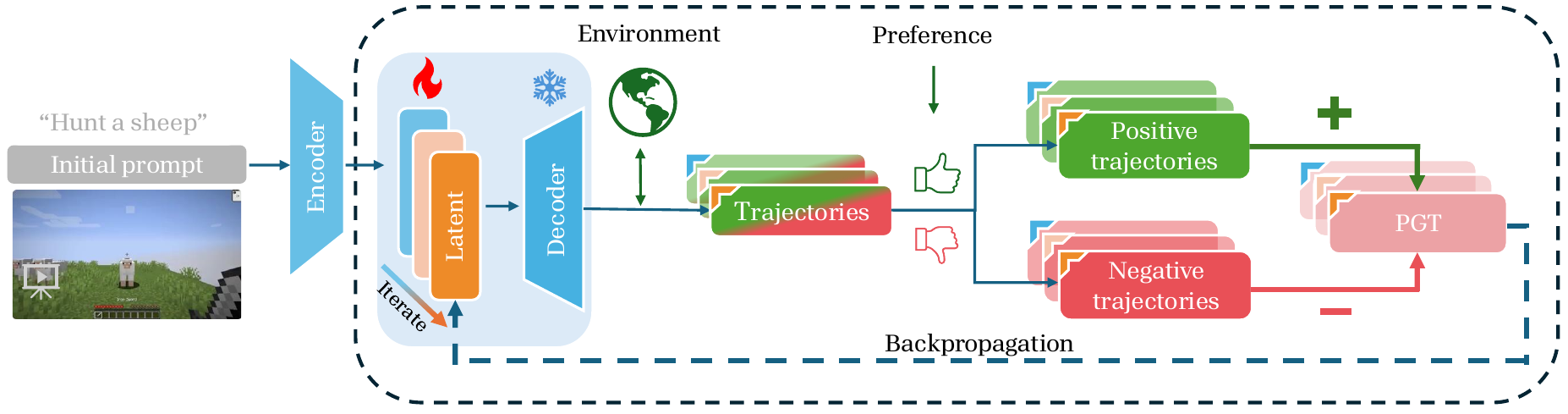}
    \caption{\textbf{Overview of the Preference Goal Tuning (PGT) framework.}
    Given a pre-trained goal-conditioned policy with frozen parameters, PGT adapts behavior solely through the latent goal.
    Starting from an initial prompt, a latent goal is inferred and used to collect trajectories.
    These trajectories are labeled with relative preferences, either by humans or by reward-based proxies.
    The latent goal is then updated via preference optimization, while the policy backbone remains fixed.
    This procedure naturally supports iterative refinement.}
    \label{fig:main_pipeline}
\end{figure*}

In this section, we introduce \emph{\textbf{P}reference \textbf{G}oal-\textbf{T}uning} (\alg), 
a post-training framework for adapting pre-trained goal-conditioned policies through preference learning over the latent goal space.
Unlike standard policy fine-tuning, \alg enforces a structural constraint: \emph{the policy parameters are frozen}, and all adaptation is realized exclusively through the latent goal representation.
An overview of the framework is shown in Figure~\ref{fig:main_pipeline}.

PGT consists of two phases:
(i) \textbf{preference sample generation}, where trajectories are collected and labeled according to preferences,
and (ii) \textbf{behavior preference propagation}, where the latent goal is updated to favor preferred behaviors.

\textbf{Problem Formulation.\ \ }
We consider a pre-trained goal-conditioned policy $\pi(a \mid s, g)$, where $g \in \mathbb{R}^d$ denotes a latent goal embedding.
The policy parameters are fixed throughout post-training.
This setting reflects practical constraints of foundation policies, where large-scale pre-training captures broad generalization and compositional knowledge that should not be overwritten by limited downstream supervision.

Under this constraint, behavioral adaptation must be achieved solely by adjusting the latent goal $g$, which serves as the designated control interface of the policy.
Rather than learning a new policy or a reward function, our objective is to identify a latent goal that induces trajectories aligned with task-specific preferences, while remaining close to the original goal representation inferred from the initial prompt.

By restricting optimization to a low-dimensional and semantically structured latent space, \alg imposes a strong inductive bias that favors generalization over memorization, which is particularly important when only a small number of preference-labeled trajectories are available.

\textbf{Preference Sample Generation.\ \ }
PGT begins with an initial prompt, which may be suboptimal, and encodes it into a latent representation $g$.
Conditioned on $g$, the frozen policy $\pi(a \mid s, g)$ interacts with the environment to generate a set of trajectories, typically on the order of $\sim 10^2$.

These trajectories are then labeled according to relative preferences.
When human supervision is available, annotators label each trajectory as either positive (preferred) or negative (non-preferred).
Given the modest number of trajectories required, the annotation cost remains manageable.

Alternatively, in environments equipped with reward functions, such as Minecraft tasks including \texttt{collect\_wood}(\includegraphics[scale=0.035,valign=c]{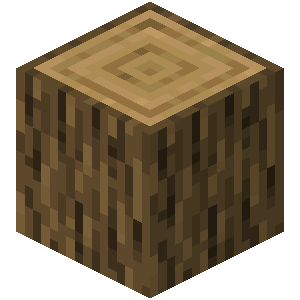}),
\texttt{tool\_bow}(\includegraphics[scale=0.6,valign=c]{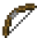}),
and \texttt{explore\_chest}(\includegraphics[scale=0.10,valign=c]{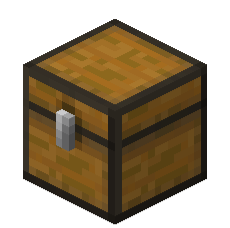}),
we use reward signals as a proxy for preference supervision.
Specifically, trajectories with the highest cumulative rewards are treated as positive samples, while those with the lowest rewards are treated as negative samples.
Importantly, reward signals are used only to induce \emph{relative preferences} between trajectories, rather than as training targets or value estimates.

\textbf{Behavior Preference Propagation.\ \ }
Given a set of positive and negative trajectories, the goal of behavior preference propagation is to update the latent goal representation such that preferred behaviors are encouraged and undesirable behaviors are suppressed, while keeping the policy network fixed.

A logical starting point is to perform behavior cloning (BC) using only positive trajectories.
However, as shown empirically in Table~\ref{tab:BC}, this approach often fails to yield consistent performance improvements.
Positive-only supervision lacks an explicit mechanism to penalize dominant but suboptimal behavior modes that are frequently encountered during rollouts, resulting in limited or unstable gains.

To solve this latent control problem without requiring a dense reward function, we leverage the probabilistic duality between control and inference. We posit that the optimal latent goal $g^*$ should induce a trajectory distribution $\pi(\tau|g)$ that minimizes the divergence from a target distribution defined by human preferences.

As derived in Appendix~\ref{sec:math}, this formulation allows us to bypass explicit reward modeling. By substituting the optimal policy form into the divergence minimization objective, we obtain a direct optimization procedure over the latent space. Specifically, given pairwise preference tuples $(\tau^{(w)}, \tau^{(l)})$, optimizing the latent goal to satisfy these preferences is mathematically equivalent to maximizing the margin between the trajectory likelihoods under the current goal $g$ and a reference goal $g_{\text{ref}}$. This yields the following tractable trajectory-wise objective:

\begin{gather}
\mathcal{L}(g, g_{\text{ref}}) =
\mathbb{E}_{(\tau^{(w)}, \tau^{(l)}) \sim \mathcal{D}}
\left[-\log \sigma(\beta \Delta)\right], \nonumber \\
\Delta =
\sum_{i=0}^T
\log \frac{\pi(a_i^{(w)} | s_i^{(w)}, g)}{\pi(a_i^{(w)} | s_i^{(w)}, g_{\text{ref}})}
-
\log \frac{\pi(a_i^{(l)} | s_i^{(l)}, g)}{\pi(a_i^{(l)} | s_i^{(l)}, g_{\text{ref}})}.
\label{eq:pgt_loss}
\end{gather}

Here, $g_{\text{ref}}$ denotes a reference latent goal, and $\beta$ controls the strength of the preference signal.
Crucially, optimization is performed solely over the latent goal $g$, while the policy $\pi$ remains fixed.
Other preference learning objectives, such as SLiC-HF~\citep{slic, slichf} and IPO~\citep{ipo}, can be readily incorporated within the same framework. The derivation is shown in Appendix~\ref{sec:math} and the result is shown in Appendix~\ref{sec:ipo}.

Restricting optimization to the latent goal offers two key advantages.
First, the latent goal serves as a semantically meaningful control interface, making it a natural locus for behavior adaptation.
Second, given the limited amount of preference data, full policy fine-tuning is highly susceptible to overfitting environment-specific details.
For example, in the Minecraft task \texttt{collect\_wood}(\includegraphics[scale=0.035,valign=c]{sections/assets/minecraft/log.png}), the agent must collect logs across diverse biomes, seeds, and initial configurations.
Full fine-tuning often memorizes spurious environment patterns, leading to degraded out-of-distribution generalization, whereas latent-only optimization preserves the robustness of the pre-trained policy.

\begin{algorithm}
\label{algo:pgt}
    \caption{Iterative Training \alg}
    \begin{algorithmic}[1]
        \State \textbf{Input:} Policy $\pi$, Number of iterations $\mathcal N$, Number of samples $\mathcal S$, Initial latent goal $g_0$, Preference learning algorithm $\mathcal A$, Maximum training epoch E
        \State \textbf{Output:} latent goal $g_\mathcal N$
        \For{iteration $i \gets 1$ to $\mathcal N$}
            \State Set $g_\text{ref}\gets g_{i-1}$
            \State Generate trajectories with $\pi$ and $g_\text{ref}$, choose the best and the worst $S$ ones into $\left\{\tau^{(w)}_s\right\}_{s=1}^{\mathcal S}$ and $\left\{\tau^{(l)}_s\right\}_{s=1}^{\mathcal S}$
            \For{epoch $e \gets 1$ to $E$}
            \State Shuffle and combine $\left\{(\tau^{(w)}_s, \tau^{(l)}_s)\right\}_{s=1}^{\mathcal S}$
            \State Optimize $g_i$ with $\mathcal A$
            \EndFor
        \EndFor
        \State \textbf{return} $g_N$
    \end{algorithmic}
\end{algorithm}

\textbf{Iterative Training.\ \ }
The proposed framework naturally supports iterative refinement.
In the first iteration, the initial prompt is encoded into a latent goal $g_0$.
Both $g$ and $g_{\text{ref}}$ are initialized with $g_0$, and preference optimization yields an updated latent goal $g_1$.
This updated goal is then used to recollect trajectories, from which new preference pairs are constructed.

\begin{figure*}[t]
    \centering
    \includegraphics[width=\linewidth]{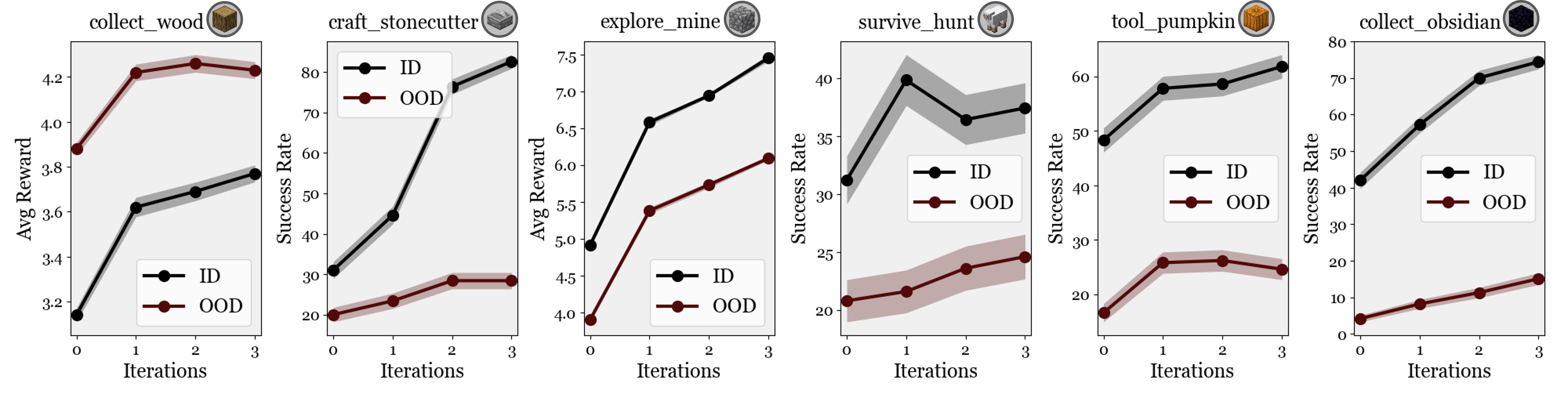}
    \caption{\textbf{Performance across iterative training rounds.}
    Each subplot corresponds to a task.
    The leftmost point represents the pre-trained policy.
    Black curves denote in-distribution evaluations, while brown curves denote out-of-distribution settings with altered initial conditions and environments.
    Performance consistently improves across iterations in both settings.}
    \label{fig:iterative}
    \vspace{-0.1 in}
\end{figure*}

By repeating this process, \alg performs successive improvement in the latent goal space under a fixed policy backbone. Empirically, we observe consistent performance gains for up to three iterations. Figure~\ref{fig:iterative} presents detailed results across multiple tasks and both in-distribution and out-of-distribution settings.

\begin{table}[ht]
\centering
\caption{\textbf{Performance improvements of \alg.} We use DPO as a representative preference-learning algorithm in the \alg framework. Both the version that trains the latent goal only and the full fine-tuning variant are implemented, demonstrating that BC is not well-suited for this setting.}
\label{tab:BC}
\resizebox{\linewidth}{!}{
\renewcommand\arraystretch{1.2}
\begin{tabular}{@{}cccclccc@{}}
\toprule
\multirow{2}{*}{Task} & \multicolumn{3}{c}{Latent-goal-only} &  & \multicolumn{3}{c}{Full Fine-Tuning} \\ \cmidrule(lr){2-4} \cmidrule(l){6-8} 
 & Pretained & BC & DPO &  & Pretained & BC & DPO \\ \midrule
\includegraphics[scale=0.043,valign=c]{sections/assets/minecraft/log.png} & 3.14 & 3.28 & \textbf{3.62} &  & 3.14 & 3.26 & \textbf{3.46} \\
\includegraphics[scale=0.7,valign=c]{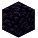} & 42.0 & 18.2 & \textbf{57.2} &  & 42.0 & 15.0 & \textbf{62.2} \\
\includegraphics[scale=0.06,valign=c]{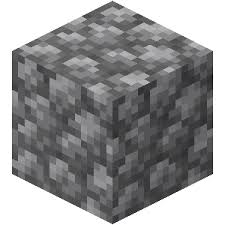} & 4.91 & 4.76 & \textbf{6.58} &  & 4.91 & 4.80 & \textbf{6.00} \\
\includegraphics[scale=0.043,valign=c]{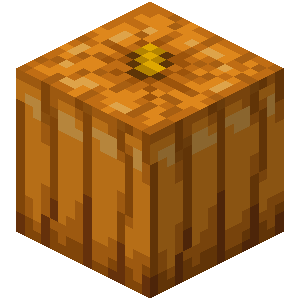} & 48.3 & 45.4 & \textbf{57.8} &  & 48.3 & 48.6 & \textbf{58.4} \\ \bottomrule
\end{tabular}}
\end{table}

\begin{figure*}[t]
    \centering
    \includegraphics[width=\linewidth]{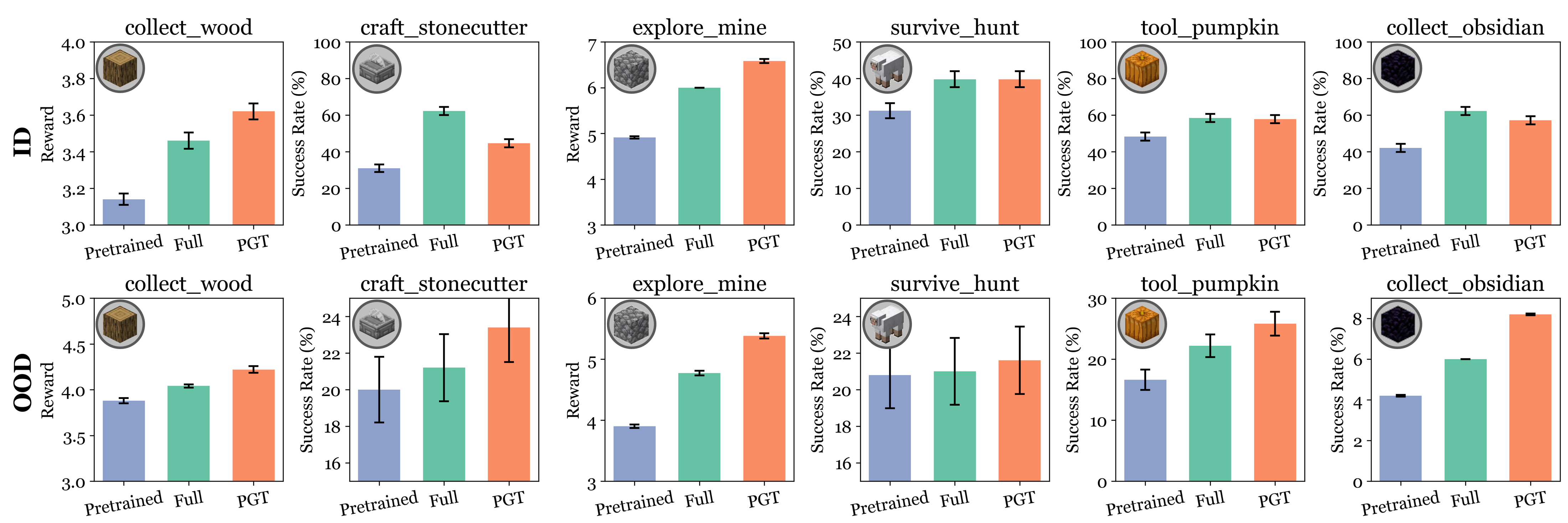}
    \caption{
        \textbf{Comparison between full fine-tuning and \alg.}
        The top row shows in-distribution (ID) performance, while the bottom row shows out-of-distribution (OOD) performance.
        Full fine-tuning improves in-distribution performance but often degrades generalization, whereas \alg achieves consistent gains in both settings. The definition of the both settings and the evaluation method are described in Appendix~\ref{OOD} and \ref{sec:hp}.
    }
    \label{fig:fpft}
    \vspace{-0.1 in}
\end{figure*}

\section{Experiments}

We evaluate \alg on Minecraft~\citep{minedojo, minestudio}, a large-scale open-ended environment that poses substantial challenges for goal-conditioned policies, including long-horizon reasoning, partial observability, and strong sensitivity to initialization and environment variations.
We select tasks from the \emph{Minecraft SkillForge} benchmark~\citep{groot}, which comprises over 30 representative skills spanning 6 major categories.
Additional details of the benchmark are provided in Appendix~\ref{skillforge}.

Our experiments are to answer the following questions:
\begin{itemize}[nosep,leftmargin=10pt]
    \item Can preference-based latent goal tuning improve both in-distribution performance and out-of-distribution generalization beyond careful prompt selection?
    \item Does isolating task-specific adaptation to compact latent goal representations mitigate task interference and preserve generalization under sequential task acquisition?
    \item Can learned latent goals serve as a robust interface between high-level planners and low-level controllers in long-horizon tasks?
\end{itemize}

Across all experiments, we report results under both in-distribution (ID) and out-of-distribution (OOD) settings.
OOD settings involve changes in environment seeds, initial conditions, and spatial configurations, while preserving the underlying task semantics.
Detailed descriptions of these settings are provided in Appendix~\ref{OOD}. Additionally, we fine-tune OpenVLA~\citep{openvla} on LIBERO-goal~\citep{libero} and compare it with other RL- or preference-based post-training methods in Section~\ref{sec:libero}, demonstrating the cross-domain applicability of \alg as a post-training framework.

\begin{table*}[t]
\centering
\caption{\textbf{Performance of different methods on tasks in \emph{Minecraft SkillForge}.} $\Delta$ represents the relative improvements in performance between policy before and after post-training (represented using ``+''). Task \texttt{collect\_wood} (\includegraphics[scale=0.035,valign=c]{sections/assets/minecraft/log.png}), \texttt{collect\_dirt} (\includegraphics[scale=0.035,valign=c]{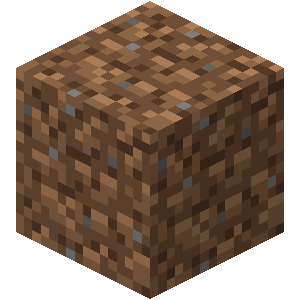}) and \texttt{survive\_plant} (\includegraphics[scale=0.035,valign=c]{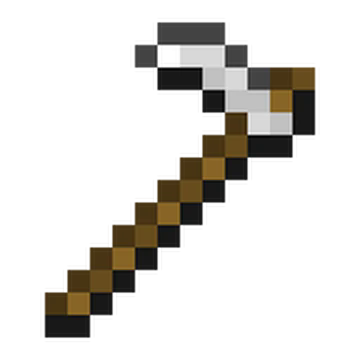}) are evaluated by the collected reward (retain one decimal place), while others are expressed as success rate and the percentage sign (\%) is omitted (retain two decimal places); for each task, the maximum time step for testing is 200; the same applies to other parts of this paper. More details are in Appendix~\ref{sec:hp}}
\label{tab:skillforge}
\resizebox{1.0\linewidth}{!}{
\renewcommand\arraystretch{1.2}
\begin{tabular}{@{}cccccccccccccc@{}}
\toprule
 & \multicolumn{6}{c}{In Distribution} &  & \multicolumn{6}{c}{Out of Distribution} \\ \cmidrule(lr){2-7} \cmidrule(l){9-14} 
\multirow{-2}{*}{Task} & GROOT & GROOT+ & $\Delta$ & STEVE & STEVE+ & $\Delta$ &  & GROOT & GROOT+ & $\Delta$ & STEVE & STEVE+ & $\Delta$ \\ \midrule
\includegraphics[scale=0.043,valign=c]{sections/assets/minecraft/log.png} & 3.14 & \textbf{3.62} & {\color[HTML]{32CB00} 15.3\%} & 3.73 & \textbf{3.90} & {\color[HTML]{32CB00} 4.6\%} &  & 3.88 & \textbf{4.22} & {\color[HTML]{32CB00} 8.8\%} & 4.22 & \textbf{4.29} & {\color[HTML]{32CB00} 1.7\%} \\
\includegraphics[scale=0.043,valign=c]{sections/assets/minecraft/dirt.png} & 27.0 & \textbf{62.8} & {\color[HTML]{32CB00} 132.6\%} & 16.3 & \textbf{36.4} & {\color[HTML]{32CB00} 123.3\%} &  & 15.4 & \textbf{54.6} & {\color[HTML]{32CB00} 254.5\%} & 30.4 & \textbf{48.0} & {\color[HTML]{32CB00} 57.9\%} \\
\includegraphics[scale=0.043,valign=c]{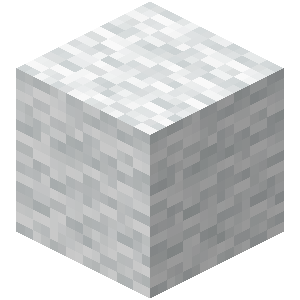} & 30.4 & \textbf{40.8} & {\color[HTML]{32CB00} 34.2\%} & 43.3 & \textbf{56.6} & {\color[HTML]{32CB00} 30.7\%} &  & 34.0 & \textbf{41.6} & {\color[HTML]{32CB00} 22.4\%} & 45.6 & \textbf{60.2} & {\color[HTML]{32CB00} 32.0\%} \\
\includegraphics[scale=0.045,valign=c]{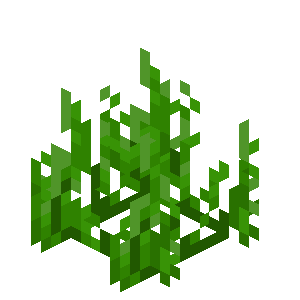} & 20.2 & \textbf{20.8} & {\color[HTML]{32CB00} 3.0\%} & 4.2 & \textbf{21.8} & {\color[HTML]{32CB00} 419.0\%} &  & 7.8 & \textbf{9.4} & {\color[HTML]{32CB00} 20.5\%} & 41.4 & \textbf{49.0} & {\color[HTML]{32CB00} 18.4\%} \\
\includegraphics[scale=0.045,valign=c]{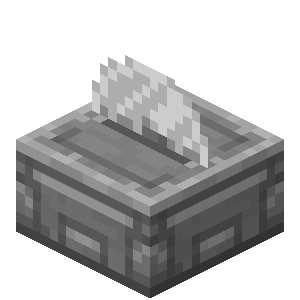} & 31.0 & \textbf{44.6} & {\color[HTML]{32CB00} 43.9\%} & 14.1 & \textbf{19.0} & {\color[HTML]{32CB00} 34.8\%} &  & 20.0 & \textbf{23.4} & {\color[HTML]{32CB00} 17.0\%} & 36.2 & \textbf{48.4} & {\color[HTML]{32CB00} 33.7\%} \\
\includegraphics[scale=0.07,valign=c]{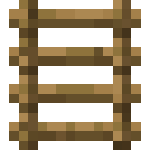} & 5.4 & \textbf{10.4} & {\color[HTML]{32CB00} 92.6\%} & 30.9 & \textbf{40.2} & {\color[HTML]{32CB00} 30.1\%} &  & 4.4 & \textbf{9.6} & {\color[HTML]{32CB00} 118.2\%} & 29.6 & \textbf{41.2} & {\color[HTML]{32CB00} 39.2\%} \\
\includegraphics[scale=0.68,valign=c]{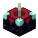} & 15.0 & \textbf{18.4} & {\color[HTML]{32CB00} 22.7\%} & 0 & 0 & - &  & 19.4 & \textbf{21.8} & {\color[HTML]{32CB00} 12.4\%} & 0 & 0 & - \\
\includegraphics[scale=0.055,valign=c]{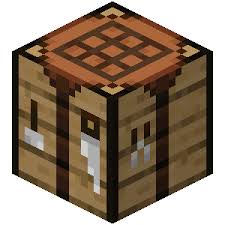} & 5.4 & \textbf{14.6} & {\color[HTML]{32CB00} 170.4\%} & 4.0 & \textbf{9.6} & {\color[HTML]{32CB00} 140.0\%} &  & 6.0 & \textbf{18.4} & {\color[HTML]{32CB00} 206.7\%} & 2.0 & \textbf{6.4} & {\color[HTML]{32CB00} 220.0\%} \\
\includegraphics[scale=0.06,valign=c]{sections/assets/minecraft/stone.jpg} & 4.91 & \textbf{6.58} & {\color[HTML]{32CB00} 34.0\%} & 6.46 & \textbf{7.32} & {\color[HTML]{32CB00} 13.3\%} &  & 3.90 & \textbf{5.38} & {\color[HTML]{32CB00} 37.9\%} & 3.49 & \textbf{5.37} & {\color[HTML]{32CB00} 53.9\%} \\
\includegraphics[scale=0.125,valign=c]{sections/assets/minecraft/chest.png} & 15.7 & \textbf{21.2} & {\color[HTML]{32CB00} 35.0\%} & 3.4 & \textbf{4.2} & {\color[HTML]{32CB00} 23.5\%} &  & \textbf{38.4} & 38.2 & {\color[HTML]{FE0000} -0.5\%} & 0.5 & \textbf{0.6} & {\color[HTML]{32CB00} 20.0\%} \\
\includegraphics[scale=0.055,valign=c]{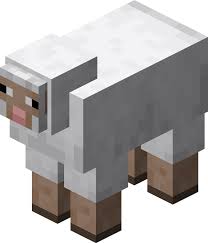} & 31.2 & \textbf{39.8} & {\color[HTML]{32CB00} 27.6\%} & \textbf{2.9} & 1.0 & {\color[HTML]{FE0000} -65.5\%} &  & 20.8 & \textbf{21.6} & {\color[HTML]{32CB00} 3.8\%} & \textbf{1} & 0.2 & {\color[HTML]{FE0000} -80.0\%} \\
\includegraphics[scale=0.055,valign=c]{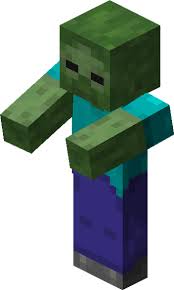} & 31.7 & \textbf{36.6} & {\color[HTML]{32CB00} 15.5\%} & 0 & 0 & - &  & 83.4 & \textbf{85.6} & {\color[HTML]{32CB00} 2.6\%} & 0 & 0 & - \\
\includegraphics[scale=0.043,valign=c]{sections/assets/minecraft/hoe.png} & 2.71 & \textbf{3.09} & {\color[HTML]{32CB00} 14.0\%} & 1.74 & \textbf{1.81} & {\color[HTML]{32CB00} 4.0\%} &  & 2.85 & \textbf{3.11} & {\color[HTML]{32CB00} 9.1\%} & 1.79 & \textbf{1.94} & {\color[HTML]{32CB00} 8.4\%} \\
\includegraphics[scale=0.043,valign=c]{sections/assets/minecraft/pumpkin.png} & 48.3 & \textbf{57.8} & {\color[HTML]{32CB00} 19.7\%} & 1.3 & \textbf{6.2} & {\color[HTML]{32CB00} 376.9\%} &  & 16.6 & \textbf{25.8} & {\color[HTML]{32CB00} 55.4\%} & 7.6 & \textbf{14.0} & {\color[HTML]{32CB00} 84.2\%} \\
\includegraphics[scale=0.6,valign=c]{sections/assets/minecraft/bow.png} & 77.4 & \textbf{85.8} & {\color[HTML]{32CB00} 10.9\%} & 88.9 & \textbf{97.8} & {\color[HTML]{32CB00} 10.0\%} &  & 77.4 & \textbf{90.6} & {\color[HTML]{32CB00} 17.1\%} & 65.2 & \textbf{88.0} & {\color[HTML]{32CB00} 35.0\%} \\
\includegraphics[scale=0.09,valign=c]{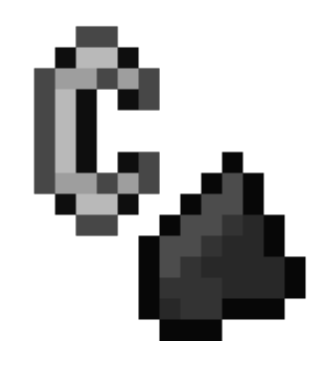} & 1.2 & \textbf{7.4} & {\color[HTML]{32CB00} 516.7\%} & 73.6 & \textbf{76.6} & {\color[HTML]{32CB00} 4.1\%} &  & 1.2 & \textbf{5.8} & {\color[HTML]{32CB00} 383.3\%} & 48.0 & \textbf{52.0} & {\color[HTML]{32CB00} 8.3\%} \\
\includegraphics[scale=0.7,valign=c]{sections/assets/minecraft/obsidian.png} & 42.0 & \textbf{57.2} & {\color[HTML]{32CB00} 36.2\%} & 0.4 & \textbf{0.7} & {\color[HTML]{32CB00} 75.0\%} &  & 4.2 & \textbf{8.2} & {\color[HTML]{32CB00} 95.2\%} & 0 & 0 & - \\ \bottomrule
\end{tabular}}
\end{table*}


\subsection{Improvement Beyond Prompt Engineering}

A natural baseline for adapting foundation policies is prompt engineering, where behavior is shaped by manually selecting or refining task descriptions.
This experiment evaluates whether \alg provides systematic improvements beyond such initialization choices, or merely compensates for suboptimal prompts.

We discard tasks in \emph{Minecraft SkillForge} that are either too difficult (zero success rate) or trivial (100\% success rate with uninformative reward).
Task selection details are described in Appendix~\ref{taskselection}.
In addition, we evaluate \texttt{explore\_climb}(\includegraphics[scale=0.05,valign=c]{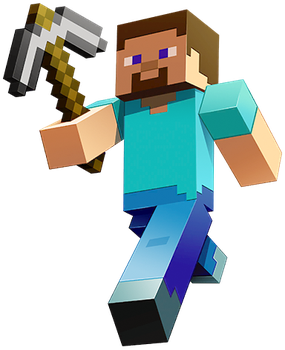}) on GROOT as a representative subjective task; details are provided in Appendix~\ref{climb}.

We experiment with two foundation policies, GROOT and STEVE-1, and apply DPO as a representative preference learning algorithm within \alg.
Results for other preference learning objectives are reported in Appendix~\ref{sec:ipo}.
For in-distribution settings, \alg achieves average relative improvements of $72.0\%$ on GROOT and $81.6\%$ on STEVE-1.
Under out-of-distribution conditions, the corresponding improvements are $73.8\%$ and $36.9\%$, respectively.
Performance gains are observed consistently across nearly all 17 evaluated tasks, with particularly notable improvements on tasks such as \texttt{collect\_dirt}(\includegraphics[scale=0.04,valign=c]{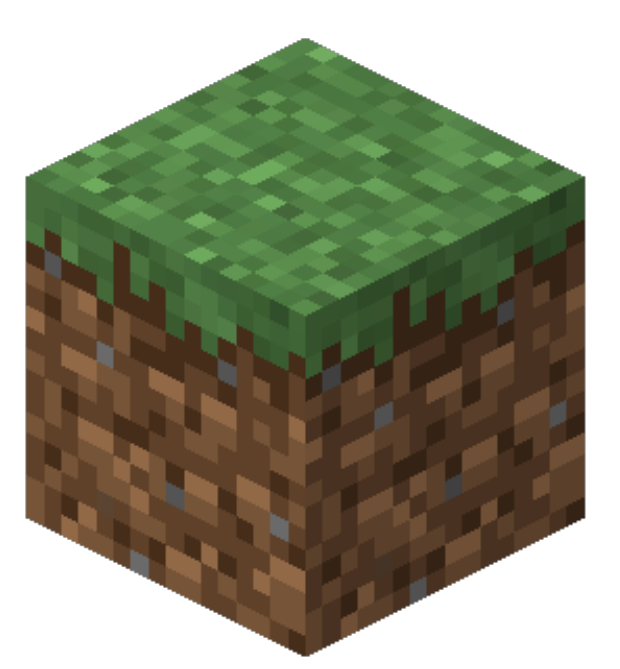}),
\texttt{craft\_crafting\_table}(\includegraphics[scale=0.05,valign=c]{sections/assets/minecraft/table.jpeg}),
and \texttt{tool\_flint}(\includegraphics[scale=0.07,valign=c]{sections/assets/minecraft/Flint_and_Steel.pdf}).
Detailed per-task results are reported in Table~\ref{tab:skillforge}.

To further examine whether the improvements from \alg depend on prompt initialization quality, we evaluate five distinct initial prompts on the representative task \texttt{collect\_wood}(\includegraphics[scale=0.035,valign=c]{sections/assets/minecraft/log.png}).
For each prompt, we perform iterative preference-based training.

\begin{figure*}[t]
    \centering
    \includegraphics[width=\linewidth]{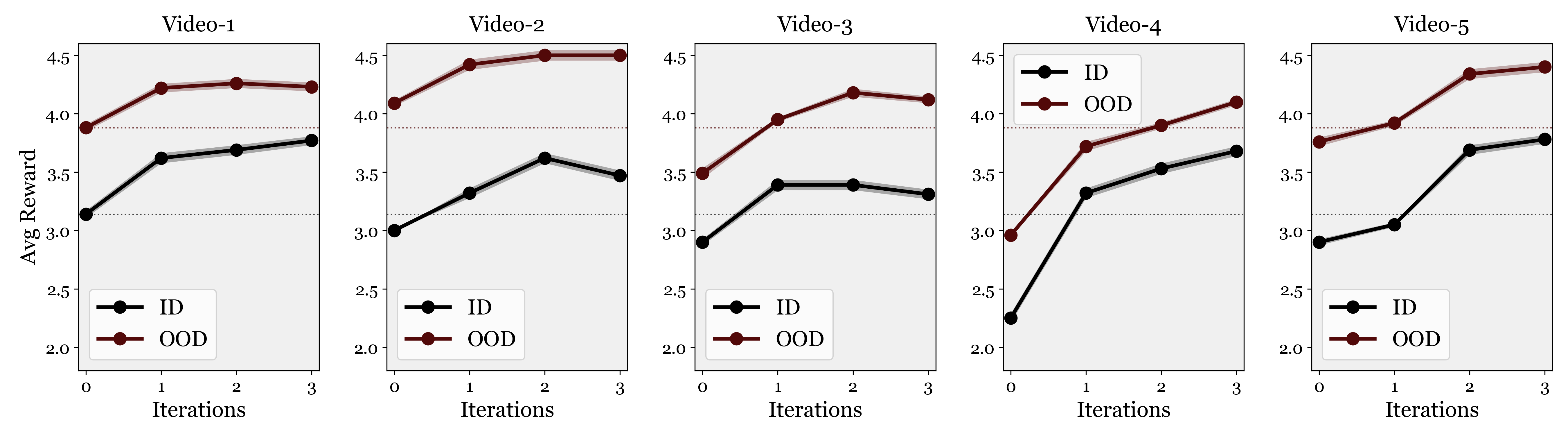}
    \caption{
        Performance under different initial prompts. Each curve corresponds to a distinct prompt, while the horizontal line denotes the best human-selected prompt.
    }
    \label{fig:diffinit}
    \vspace{-0.1 in}
\end{figure*}

As shown in Figure~\ref{fig:diffinit}, regardless of the initial prompt—including those already yielding strong baseline performance—the optimized goals consistently outperform the best human-selected reference prompt.
This result indicates that the gains achieved by \alg cannot be attributed to prompt engineering alone, but arise from preference-driven reshaping of the induced behavior distribution.

\begin{table*}[t]
\centering
\caption{Multi-task learning on 6 Minecraft tasks. }
\label{tab:MTL}
\resizebox{0.68\linewidth}{!}{
\renewcommand\arraystretch{1.2}
\begin{tabular}{@{}cccccccccc@{}}
\toprule
\multirow{2}{*}{Task} & \multicolumn{4}{c}{In Distribution(ID)} &  & \multicolumn{4}{c}{Out of Distribution(OOD)} \\ \cmidrule(lr){2-5} \cmidrule(l){7-10} 
 & Pretrained & Ensemble & MTL & \alg(Ours) &  & Pretrained & Ensemble & MTL & \alg(Ours) \\ \midrule
\includegraphics[scale=0.043,valign=c]{sections/assets/minecraft/log.png} & 3.14 & 3.46 & \textbf{3.64} & \underline{3.62} &  & 3.88 & 4.04 & \textbf{4.30} & \underline{4.22} \\
\includegraphics[scale=0.045,valign=c]{sections/assets/minecraft/stonecutter.png} & 31.0 & \underline{62.2} & \textbf{66.8} & 44.6 &  & 20.0 & \underline{21.2} & 18.6 & \textbf{23.4} \\
\includegraphics[scale=0.06,valign=c]{sections/assets/minecraft/stone.jpg} & 4.91 & \underline{6.00} & 5.98 & \textbf{6.58} &  & 3.90 & \underline{4.77} & 4.70 & \textbf{5.38} \\
\includegraphics[scale=0.05,valign=c]{sections/assets/minecraft/hunt.jpg} & 31.2 & \underline{39.8} & \textbf{44.2} & \underline{39.8} &  & 20.8 & 21.0 & \textbf{31.4} & \underline{21.6} \\
\includegraphics[scale=0.043,valign=c]{sections/assets/minecraft/pumpkin.png} & 48.3 & \underline{58.4} & \textbf{61.4} & 57.8 &  & 16.6 & 22.2 & \underline{22.8} & \textbf{25.8} \\
\includegraphics[scale=0.7,valign=c]{sections/assets/minecraft/obsidian.png} & 42.0 & \textbf{62.2} & 53.2 & \underline{57.2} &  & 4.2 & 6.0 & \textbf{10.2} & \underline{8.2} \\ \bottomrule
\end{tabular}}
\end{table*}
\begin{table*}[]
\centering
\caption{Different continual learning baselines.}
\label{tab:CL}
\resizebox{0.65\linewidth}{!}{
\renewcommand\arraystretch{1.2}
\begin{tabular}{cccccccccccc}
\toprule
\multirow{2}{*}{Task} & \multicolumn{5}{c}{In Distribution(ID)} &  & \multicolumn{5}{c}{Out of Distribution(OOD)} \\ \cline{2-6} \cline{8-12} 
 & ER & EWC & KD & NCL & \alg(Ours) &  & ER & EWC & KD & NCL & \alg(Ours) \\ \cline{1-6} \cline{8-12} 
\includegraphics[scale=0.7,valign=c]{sections/assets/minecraft/obsidian.png} & 60.2 & \underline{64.6} & \textbf{66.8} & 61.2 & 57.2 &  & 6.0 & 5.4 & 5.4 & \underline{6.8} & \textbf{8.2} \\
\includegraphics[scale=0.043,valign=c]{sections/assets/minecraft/pumpkin.png} & \textbf{65.4} & 60.0 & 60.8 & \underline{61.4} & 57.8 &  & \underline{25.0} & 23.8 & 20.6 & 20.4 & \textbf{25.8} \\
\includegraphics[scale=0.055,valign=c]{sections/assets/minecraft/table.jpeg} & \underline{8.6} & 6.8 & 6.8 & 7.2 & \textbf{14.6} &  & \underline{9.0} & 7.4 & 5.8 & 7.0 & \textbf{18.4} \\ \bottomrule
\end{tabular}}
\end{table*}



\subsection{Sequential Task Adaptation}

We next evaluate \alg under a sequential task acquisition setting, where tasks are learned one after another without revisiting earlier training data.
In contrast to prior approaches that adapt a single shared set of policy parameters across tasks, \alg performs task-specific adaptation exclusively in the latent goal space, storing a compact latent representation for each task. As a result, task interference is avoided by construction, rather than mitigated through regularization, replay, or consolidation mechanisms. We compare \alg with multi-task learning (MTL) and several standard continual learning baselines: naive continual learning (NCL), knowledge distillation (KD)~\citep{kd}, experience replay (ER)~\citep{er}, and elastic weight consolidation (EWC)~\citep{ewc}. All of them are implemented using full fine-tuning.

We first evaluate MTL on six representative tasks. As shown in Table~\ref{tab:MTL}, while \alg performs marginally worse than MTL in in-distribution scenarios, it achieves better aggregate performance in out-of-distribution settings, indicating improved generalization via latent goal space adaptation.

We further conduct sequential continual learning in the following order:
\texttt{collect\_obsidian}(\includegraphics[scale=0.6,valign=c]{sections/assets/minecraft/obsidian.png}) $\to$
\texttt{tool\_pumpkin}(\includegraphics[scale=0.04,valign=c]{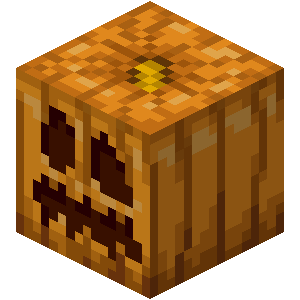}) $\to$
\texttt{craft\_crafting\_table}(\includegraphics[scale=0.05,valign=c]{sections/assets/minecraft/table.jpeg}) $\to$
\texttt{explore\_climb}(\includegraphics[scale=0.035,valign=c]{sections/assets/minecraft/Steve.png}).

The performance after learning all four tasks is reported in Table~\ref{tab:CL}, with intermediate results provided in Appendix~\ref{CL}.
Results for NCL, KD, ER, and EWC are shown in Tables~\ref{tab:NCL}–\ref{tab:EWC}.

Overall, \alg demonstrates strong robustness across tasks and environments, achieving superior out-of-distribution generalization while avoiding catastrophic forgetting.
These results indicate that the advantage of \alg stems not from improved consolidation strategies, but from avoiding parameter interference altogether through latent goal isolation.
While \alg does not aim to solve continual learning in the classical sense of a single shared parameterization, these results demonstrate that isolating task adaptation to latent goals effectively addresses several core challenges commonly encountered in continual learning, including catastrophic forgetting and negative transfer.

\begin{figure*}[t]
    \centering
    \includegraphics[width=\linewidth]{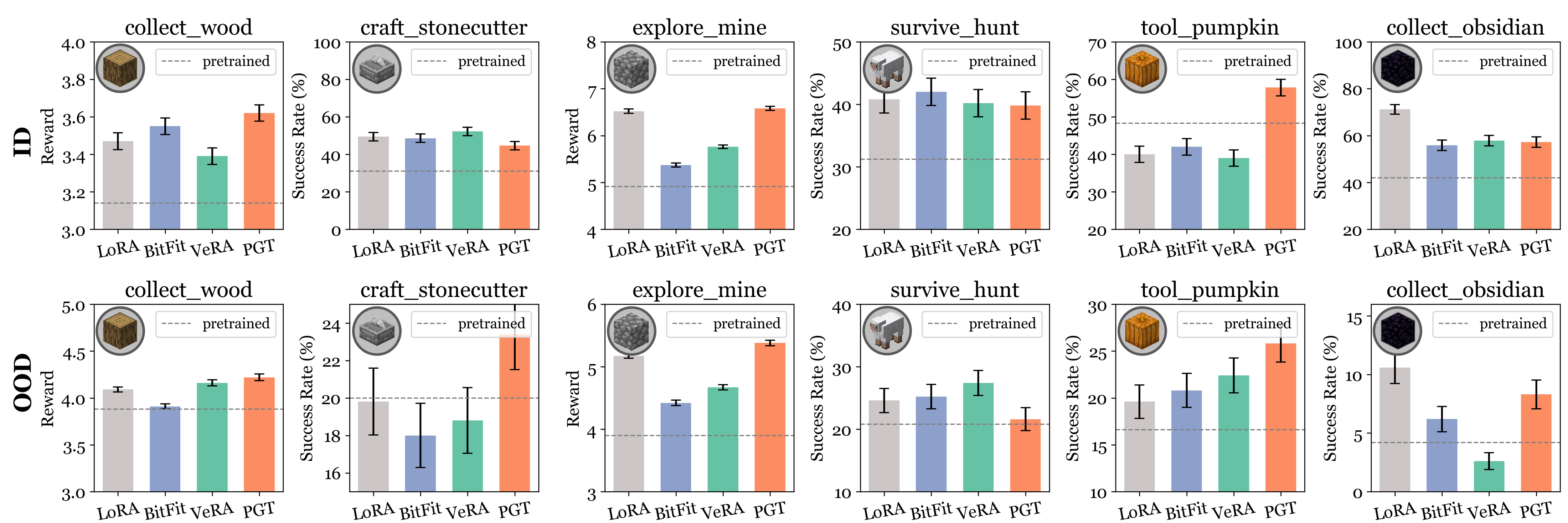}
    \caption{
        \textbf{Comparison of parameter-efficient fine-tuning (PEFT) methods.} The horizontal line denotes the performance of the pretrained GROOT model. The upper section reports results under in-distribution settings, while the lower section reports results under out-of-distribution settings. \alg demonstrates competitive and often superior performance across most settings.
    }
    \label{fig:pe}
\end{figure*}

\subsection{Long-Horizon Tasks with Planner–Controller Decomposition}

We further evaluate \alg in a long-horizon setting by combining a high-level planner with a low-level controller.
Specifically, we integrate the GROOT agent with the JARVIS-1 planner~\citep{jarvis1} to perform item crafting tasks from scratch in a forest environment with random initial orientations.

The planner generates high-level action sequences, while the controller executes them under the guidance of the learned goal.
Agents are allowed 1000 timesteps, and we evaluate five representative items along the wood-related technology tree.
Results are reported in Table~\ref{tab:longhorizon}.

Compared to the baseline, \alg consistently improves long-horizon task success.
These results suggest that \alg-trained latent goals can serve as a behavior-level alignment interface between reactive planners and visuomotor controllers, improving robustness without modifying either component.

\begin{table}[h]
\centering
\caption{\textbf{Success rates (\%) on long-horizon tasks: crafting items from scratch.} The latent goal matches that of GROOT+ in Table~\ref{tab:skillforge}. Results are evaluated over a total of 1,000 trials across 2 different seeds. The maximum steps for testing long-horizon tasks is 2000.}
\resizebox{0.8\linewidth}{!}{
\renewcommand\arraystretch{1.2}
\begin{tabular}{cccccc}
\toprule
Task & \includegraphics[scale=0.09,valign=c]{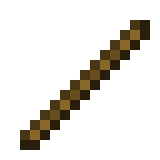} & \includegraphics[scale=0.035,valign=c]{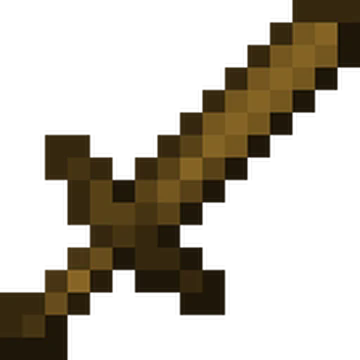} & \includegraphics[scale=0.09,valign=c]{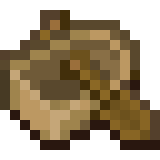} & \includegraphics[scale=0.06,valign=c]{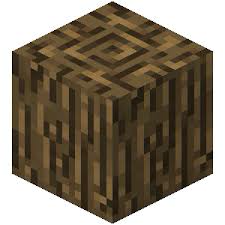} & \includegraphics[scale=0.09,valign=c]{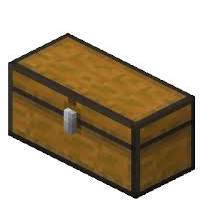} \\
\hline
Pretrained & 99.5 & 94.0 & 80.7 & 60.8 & 37.8 \\
\alg(Ours) & \textbf{100} & \textbf{100} & \textbf{89.5} & \textbf{80.7} & \textbf{64.9} \\
\bottomrule
\end{tabular}}
\label{tab:longhorizon}
\end{table}

\vspace{-0.1 in}
\subsection{Ablation Study on Parameter-Efficient Fine-Tuning}

Finally, we compare \alg with other parameter-efficient fine-tuning (PEFT) methods, including LoRA~\citep{lora}, BitFit~\citep{bitfit}, and VeRA~\citep{vera}.
For all methods, we use identical preference data and fine-tune the entire model for PEFT baselines, while \alg optimizes only the latent goal.

As shown in Figure~\ref{fig:pe}, \alg achieves competitive or superior performance under both in-distribution and out-of-distribution settings.

\subsection{Cross-Domain Validation on Robotic Manipulation}
\label{sec:libero}

\begin{figure*}[t]
    \centering
    \includegraphics[width=\linewidth]{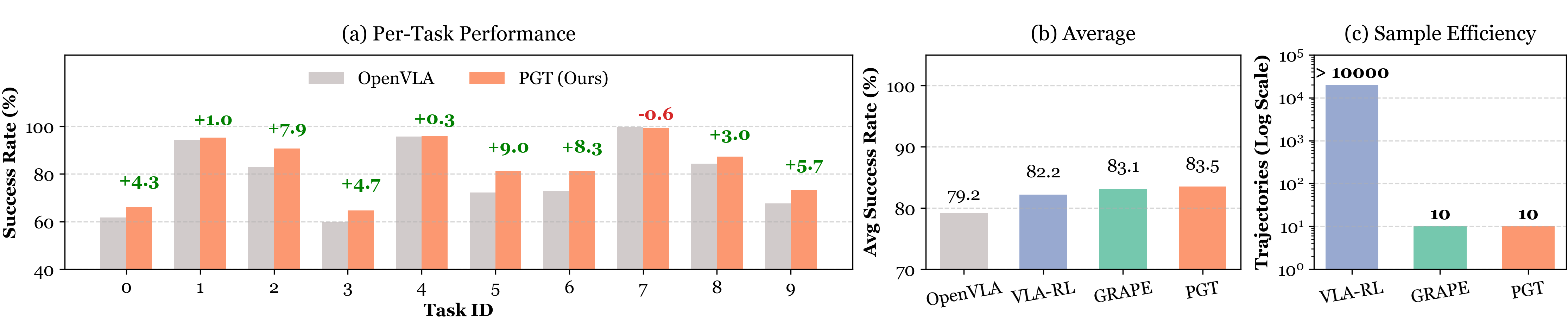}
    \caption{
        \textbf{Cross-domain validation on LIBERO-goal~\citep{libero} benchmark.} \alg delivers higher performance and sample efficiency than PPO-based VLA-RL~\citep{ppo, vlarl}, while matching GRAPE~\citep{dpo, grape} with a much simpler preference-ranking function.
    }
    \label{fig:libero}
\end{figure*}

To examine whether the effectiveness of \alg is specific to Minecraft or reflects a more general post-training principle, we further evaluate \alg in a robotic manipulation domain.
Specifically, we conduct experiments using the OpenVLA policy on the LIBERO-goal benchmark, which consists of a suite of goal-conditioned tabletop manipulation tasks with sparse success-based rewards.

LIBERO-goal differs substantially from Minecraft in state representation, action space, and task structure.
While Minecraft involves language-conditioned agents operating in discrete, open-ended environments, LIBERO-goal focuses on visuomotor control with precise object interactions.
As a result, this setting provides a strong test of whether preference-based latent goal tuning can generalize across domains with fundamentally different inductive biases. For each task, we freeze the OpenVLA~\citep{openvla} policy backbone and treat the token-level word embedding as the sole adaptation interface.
Initial goal embeddings are obtained using the default task descriptions provided by the benchmark.
During post-training, we collect rollouts conditioned on the current goal embedding and construct preference supervision based on task success, treating successful trajectories as preferred over failed ones.
We then apply the same preference optimization procedure, optimizing only the soft prompt input while keeping the Prismatic-7B backbone~\citep{prismatic} fixed.

Figure~\ref{fig:libero} summarizes the performance of \alg in LIBERO-goal benchmark~\citep{libero} compared to the OpenVLA-libero-goal~\citep{openvla}, VLA-RL~\citep{vlarl} and GRAPE~\citep{grape}. Notably, performance improvements are observed even on tasks where the baseline policy already achieves strong success rates, indicating that the benefits of \alg are not limited to compensating for poor initialization. These results suggest that preference-based optimization over the latent goal space can effectively reshape induced behavior distributions without modifying the underlying policy, even in continuous-control robotic domains.

Overall, this cross-domain evaluation demonstrates that the effectiveness of \alg is not tied to the specific characteristics of Minecraft, but extends to robotic manipulation tasks with distinct state, action, and reward structures.

\section{Conclusion and Limitations}

In this work, we presented \emph{\textbf{P}reference \textbf{G}oal-\textbf{T}uning} (\alg), a framework that reformulates post-training adaptation as a latent control problem. Instead of modifying the policy parameters, \alg optimizes a continuous latent goal to align the induced trajectory distribution with task preferences. By leveraging a theoretically grounded preference learning objective, \alg significantly enhances the capabilities of foundation policies with minimal data, consistently outperforming expert-crafted prompts on the \emph{Minecraft SkillForge} benchmark. Crucially, our experiments demonstrate that decoupling task alignment (via the latent goal) from physical dynamics (via the frozen policy) leads to superior robustness in out-of-distribution settings and effective long-horizon control when combined with high-level planners.

While \alg demonstrates remarkable effectiveness, it entails certain limitations intrinsic to its design. First, as a latent control framework, \alg operates by navigating the behavioral manifold of the frozen policy. This reliance implies that the base model must possess at least a marginal capability to execute the task; if the pretrained policy fails to sample any successful trajectories for preference construction, \alg cannot extract the necessary signal to guide optimization. Second, the optimization is task-specific: \alg learns a distinct latent control vector for each task rather than a universal generalized mapping. However, this modularity is also a strength; \alg functions as a non-destructive, plug-and-play interface that elicits optimal behaviors without altering the underlying foundation model. This ensures that the original generalization capabilities of the pretrained policy remain intact, allowing the model to handle unseen tasks seamlessly in its original zero-shot capacity.

\section*{Impact Statements}

This paper introduces the \alg framework for instruction-following policy post-training, which leverages a small amount of online synthesized data to efficiently enhance model performance, aiming to advance the field of machine learning. Our work carries several potential societal implications. For instance, the OOD generalization capability of the execution environment enables instruction-following robots operating in hazardous scenarios to utilize simulator data to improve their task execution in real-world settings. At present, we do not identify any ethical concerns that require special emphasis.

\bibliographystyle{abbrvnat}
\bibliography{reference}

@misc{groot,
    title={GROOT: Learning to Follow Instructions by Watching Gameplay Videos}, 
    author={Shaofei Cai and Bowei Zhang and Zihao Wang and Xiaojian Ma and Anji Liu and Yitao Liang},
    year={2023},
    eprint={2310.08235},
    archivePrefix={arXiv},
    primaryClass={cs.AI}
}

@article{jarvis1,
    title   = {JARVIS-1: Open-World Multi-task Agents with Memory-Augmented Multimodal Language Models},
    author  = {Zihao Wang and Shaofei Cai and Anji Liu and Yonggang Jin and Jinbing Hou and Bowei Zhang and Haowei Lin and Zhaofeng He and Zilong Zheng and Yaodong Yang and Xiaojian Ma and Yitao Liang},
    year    = {2023},
    journal = {arXiv preprint arXiv: 2311.05997}
    }

@misc{vpt,
      title={Video PreTraining (VPT): Learning to Act by Watching Unlabeled Online Videos}, 
      author={Bowen Baker and Ilge Akkaya and Peter Zhokhov and Joost Huizinga and Jie Tang and Adrien Ecoffet and Brandon Houghton and Raul Sampedro and Jeff Clune},
      year={2022},
      eprint={2206.11795},
      archivePrefix={arXiv},
      primaryClass={cs.LG},
      url={https://arxiv.org/abs/2206.11795}, 
}

@article{steve1,
  title={Steve-1: A generative model for text-to-behavior in minecraft},
  author={Lifshitz, Shalev and Paster, Keiran and Chan, Harris and Ba, Jimmy and McIlraith, Sheila},
  journal={Advances in Neural Information Processing Systems},
  volume={36},
  year={2024}
}

@inproceedings{minedojo,
  title     = {MineDojo: Building Open-Ended Embodied Agents with Internet-Scale Knowledge},
  author    = {Linxi Fan and Guanzhi Wang and Yunfan Jiang and Ajay Mandlekar and Yuncong Yang and Haoyi Zhu and Andrew Tang and De-An Huang and Yuke Zhu and Anima Anandkumar},
  booktitle = {Thirty-sixth Conference on Neural Information Processing Systems Datasets and Benchmarks Track},
  year      = {2022},
  url       = {https://openreview.net/forum?id=rc8o_j8I8PX}
}

@misc{dpo,
      title={Direct Preference Optimization: Your Language Model is Secretly a Reward Model}, 
      author={Rafael Rafailov and Archit Sharma and Eric Mitchell and Stefano Ermon and Christopher D. Manning and Chelsea Finn},
      year={2024},
      eprint={2305.18290},
      archivePrefix={arXiv},
      primaryClass={cs.LG},
      url={https://arxiv.org/abs/2305.18290}, 
}

@inproceedings{groot2,
  title={GROOT-2: Weakly Supervised Multimodal Instruction Following Agents},
  author={Cai, Shaofei and Zhang, Bowei and Wang, Zihao and Lin, Haowei and Ma, Xiaojian and Liu, Anji and Liang, Yitao},
  booktitle={The Thirteenth International Conference on Learning Representations},
  year={2025}
}

@misc{ppo,
      title={Proximal Policy Optimization Algorithms}, 
      author={John Schulman and Filip Wolski and Prafulla Dhariwal and Alec Radford and Oleg Klimov},
      year={2017},
      eprint={1707.06347},
      archivePrefix={arXiv},
      primaryClass={cs.LG},
      url={https://arxiv.org/abs/1707.06347}, 
}

@misc{lmrlhf,
      title={Fine-Tuning Language Models from Human Preferences}, 
      author={Daniel M. Ziegler and Nisan Stiennon and Jeffrey Wu and Tom B. Brown and Alec Radford and Dario Amodei and Paul Christiano and Geoffrey Irving},
      year={2020},
      eprint={1909.08593},
      archivePrefix={arXiv},
      primaryClass={cs.CL},
      url={https://arxiv.org/abs/1909.08593}, 
}

@misc{rt2,
      title={RT-2: Vision-Language-Action Models Transfer Web Knowledge to Robotic Control}, 
      author={Anthony Brohan and Noah Brown and Justice Carbajal and Yevgen Chebotar and Xi Chen and Krzysztof Choromanski and Tianli Ding and Danny Driess and Avinava Dubey and Chelsea Finn and Pete Florence and Chuyuan Fu and Montse Gonzalez Arenas and Keerthana Gopalakrishnan and Kehang Han and Karol Hausman and Alexander Herzog and Jasmine Hsu and Brian Ichter and Alex Irpan and Nikhil Joshi and Ryan Julian and Dmitry Kalashnikov and Yuheng Kuang and Isabel Leal and Lisa Lee and Tsang-Wei Edward Lee and Sergey Levine and Yao Lu and Henryk Michalewski and Igor Mordatch and Karl Pertsch and Kanishka Rao and Krista Reymann and Michael Ryoo and Grecia Salazar and Pannag Sanketi and Pierre Sermanet and Jaspiar Singh and Anikait Singh and Radu Soricut and Huong Tran and Vincent Vanhoucke and Quan Vuong and Ayzaan Wahid and Stefan Welker and Paul Wohlhart and Jialin Wu and Fei Xia and Ted Xiao and Peng Xu and Sichun Xu and Tianhe Yu and Brianna Zitkovich},
      year={2023},
      eprint={2307.15818},
      archivePrefix={arXiv},
      primaryClass={cs.RO},
      url={https://arxiv.org/abs/2307.15818}, 
}

@misc{lora,
      title={LoRA: Low-Rank Adaptation of Large Language Models}, 
      author={Edward J. Hu and Yelong Shen and Phillip Wallis and Zeyuan Allen-Zhu and Yuanzhi Li and Shean Wang and Lu Wang and Weizhu Chen},
      year={2021},
      eprint={2106.09685},
      archivePrefix={arXiv},
      primaryClass={cs.CL},
      url={https://arxiv.org/abs/2106.09685}, 
}

@misc{vera,
      title={VeRA: Vector-based Random Matrix Adaptation}, 
      author={Dawid J. Kopiczko and Tijmen Blankevoort and Yuki M. Asano},
      year={2024},
      eprint={2310.11454},
      archivePrefix={arXiv},
      primaryClass={cs.CL},
      url={https://arxiv.org/abs/2310.11454}, 
}

@misc{bitfit,
      title={BitFit: Simple Parameter-efficient Fine-tuning for Transformer-based Masked Language-models}, 
      author={Elad Ben Zaken and Shauli Ravfogel and Yoav Goldberg},
      year={2022},
      eprint={2106.10199},
      archivePrefix={arXiv},
      primaryClass={cs.LG},
      url={https://arxiv.org/abs/2106.10199}, 
}

@article{rlhf,
  title={Deep reinforcement learning from human preferences},
  author={Christiano, Paul F and Leike, Jan and Brown, Tom and Martic, Miljan and Legg, Shane and Amodei, Dario},
  journal={Advances in neural information processing systems},
  volume={30},
  year={2017}
}

@inproceedings{dualprompt,
  title={Dualprompt: Complementary prompting for rehearsal-free continual learning},
  author={Wang, Zifeng and Zhang, Zizhao and Ebrahimi, Sayna and Sun, Ruoxi and Zhang, Han and Lee, Chen-Yu and Ren, Xiaoqi and Su, Guolong and Perot, Vincent and Dy, Jennifer and others},
  booktitle={European Conference on Computer Vision},
  pages={631--648},
  year={2022},
  organization={Springer}
}

@inproceedings{ipo,
  title={A general theoretical paradigm to understand learning from human preferences},
  author={Azar, Mohammad Gheshlaghi and Guo, Zhaohan Daniel and Piot, Bilal and Munos, Remi and Rowland, Mark and Valko, Michal and Calandriello, Daniele},
  booktitle={International Conference on Artificial Intelligence and Statistics},
  pages={4447--4455},
  year={2024},
  organization={PMLR}
}

@inproceedings{malmo,
  title={The Malmo Platform for Artificial Intelligence Experimentation.},
  author={Johnson, Matthew and Hofmann, Katja and Hutton, Tim and Bignell, David},
  booktitle={Ijcai},
  volume={16},
  pages={4246--4247},
  year={2016}
}

@article{openvla,
  title={OpenVLA: An Open-Source Vision-Language-Action Model},
  author={Kim, Moo Jin and Pertsch, Karl and Karamcheti, Siddharth and Xiao, Ted and Balakrishna, Ashwin and Nair, Suraj and Rafailov, Rafael and Foster, Ethan and Lam, Grace and Sanketi, Pannag and others},
  journal={arXiv preprint arXiv:2406.09246},
  year={2024}
}

@article{ewc,
   title={Overcoming catastrophic forgetting in neural networks},
   volume={114},
   ISSN={1091-6490},
   url={http://dx.doi.org/10.1073/pnas.1611835114},
   DOI={10.1073/pnas.1611835114},
   number={13},
   journal={Proceedings of the National Academy of Sciences},
   publisher={Proceedings of the National Academy of Sciences},
   author={Kirkpatrick, James and Pascanu, Razvan and Rabinowitz, Neil and Veness, Joel and Desjardins, Guillaume and Rusu, Andrei A. and Milan, Kieran and Quan, John and Ramalho, Tiago and Grabska-Barwinska, Agnieszka and Hassabis, Demis and Clopath, Claudia and Kumaran, Dharshan and Hadsell, Raia},
   year={2017},
   month=mar, pages={3521–3526} }

@misc{er,
      title={Gradient Episodic Memory for Continual Learning}, 
      author={David Lopez-Paz and Marc'Aurelio Ranzato},
      year={2022},
      eprint={1706.08840},
      archivePrefix={arXiv},
      primaryClass={cs.LG},
      url={https://arxiv.org/abs/1706.08840}, 
}

@misc{kd,
      title={Distilling the Knowledge in a Neural Network}, 
      author={Geoffrey Hinton and Oriol Vinyals and Jeff Dean},
      year={2015},
      eprint={1503.02531},
      archivePrefix={arXiv},
      primaryClass={stat.ML},
      url={https://arxiv.org/abs/1503.02531}, 
}

@article{simpo,
  title={Simpo: Simple preference optimization with a reference-free reward},
  author={Meng, Yu and Xia, Mengzhou and Chen, Danqi},
  journal={arXiv preprint arXiv:2405.14734},
  year={2024}
}

@misc{bc-z,
      title={BC-Z: Zero-Shot Task Generalization with Robotic Imitation Learning}, 
      author={Eric Jang and Alex Irpan and Mohi Khansari and Daniel Kappler and Frederik Ebert and Corey Lynch and Sergey Levine and Chelsea Finn},
      year={2022},
      eprint={2202.02005},
      archivePrefix={arXiv},
      primaryClass={cs.RO},
      url={https://arxiv.org/abs/2202.02005}, 
}

@misc{minerl,
      title={MineRL: A Large-Scale Dataset of Minecraft Demonstrations}, 
      author={William H. Guss and Brandon Houghton and Nicholay Topin and Phillip Wang and Cayden Codel and Manuela Veloso and Ruslan Salakhutdinov},
      year={2019},
      eprint={1907.13440},
      archivePrefix={arXiv},
      primaryClass={cs.LG},
      url={https://arxiv.org/abs/1907.13440}, 
}

@inproceedings{deps,
  title={Describe, explain, plan and select: interactive planning with large language models enables open-world multi-task agents},
  author={Wang, Zihao and Cai, Shaofei and Chen, Guanzhou and Liu, Anji and Ma, Xiaojian and Liang, Yitao and CraftJarvis, Team},
  booktitle={Proceedings of the 37th International Conference on Neural Information Processing Systems},
  pages={34153--34189},
  year={2023}
}

@inproceedings{opencontrol,
  title={Open-world multi-task control through goal-aware representation learning and adaptive horizon prediction},
  author={Cai, Shaofei and Wang, Zihao and Ma, Xiaojian and Liu, Anji and Liang, Yitao},
  booktitle={Proceedings of the IEEE/CVF Conference on Computer Vision and Pattern Recognition},
  pages={13734--13744},
  year={2023}
}

@article{trueskill,
  title={TrueSkill™: a Bayesian skill rating system},
  author={Herbrich, Ralf and Minka, Tom and Graepel, Thore},
  journal={Advances in neural information processing systems},
  volume={19},
  year={2006}
}

@article{slic,
  title={Calibrating sequence likelihood improves conditional language generation},
  author={Zhao, Yao and Khalman, Misha and Joshi, Rishabh and Narayan, Shashi and Saleh, Mohammad and Liu, Peter J},
  journal={arXiv preprint arXiv:2210.00045},
  year={2022}
}

@article{slichf,
  title={Slic-hf: Sequence likelihood calibration with human feedback},
  author={Zhao, Yao and Joshi, Rishabh and Liu, Tianqi and Khalman, Misha and Saleh, Mohammad and Liu, Peter J},
  journal={arXiv preprint arXiv:2305.10425},
  year={2023}
}

@article{rlgen,
  title={A Survey of Generalisation in Deep Reinforcement Learning},
  author={Kirk, Robert and Zhang, Amy and Grefenstette, Edward and Rocktäsche, Tim},
  journal={arXiv preprint arXiv:2111.09794},
  year={2021}
}

@article{jin2023mini,
  title={Mini-BEHAVIOR: A Procedurally Generated Benchmark for Long-horizon Decision-Making in Embodied AI},
  author={Jin, Emily and Hu, Jiaheng and Huang, Zhuoyi and Zhang, Ruohan and Wu, Jiajun and Fei-Fei, Li and Mart{\'\i}n-Mart{\'\i}n, Roberto},
  journal={arXiv preprint arXiv:2310.01824},
  year={2023}
}

@article{zhang2020high,
  title={High precision control and deep learning-based corn stand counting algorithms for agricultural robot},
  author={Zhang, Zhongzhong and Kayacan, Erkan and Thompson, Benjamin and Chowdhary, Girish},
  journal={Autonomous Robots},
  volume={44},
  number={7},
  pages={1289--1302},
  year={2020},
  publisher={Springer}
}

@inproceedings{yang2023essential,
  title={What is essential for unseen goal generalization of offline goal-conditioned rl?},
  author={Yang, Rui and Yong, Lin and Ma, Xiaoteng and Hu, Hao and Zhang, Chongjie and Zhang, Tong},
  booktitle={International Conference on Machine Learning},
  pages={39543--39571},
  year={2023},
  organization={PMLR}
}

@article{pi0,
  title={$\pi_0$: A Vision-Language-Action Flow Model for General Robot Control},
  author={Black, Kevin and Brown, Noah and Driess, Danny and Esmail, Adnan and Equi, Michael and Finn, Chelsea and Fusai, Niccolo and Groom, Lachy and Hausman, Karol and Ichter, Brian and others},
  journal={arXiv preprint arXiv:2410.24164},
  year={2024}
}

@article{pi05,
  title={$\pi_{0.5}$: a Vision-Language-Action Model with Open-World Generalization},
  author={Intelligence, Physical and Black, Kevin and Brown, Noah and Darpinian, James and Dhabalia, Karan and Driess, Danny and Esmail, Adnan and Equi, Michael and Finn, Chelsea and Fusai, Niccolo and others},
  journal={arXiv preprint arXiv:2504.16054},
  year={2025}
}

@article{humantorobot,
  title={Emergence of Human to Robot Transfer in Vision-Language-Action Models},
  author={Kareer, Simar and Pertsch, Karl and Darpinian, James and Hoffman, Judy and Xu, Danfei and Levine, Sergey and Finn, Chelsea and Nair, Suraj},
  journal={arXiv preprint arXiv:2512.22414},
  year={2025}
}

@inproceedings{rocket1,
  title={Rocket-1: Mastering open-world interaction with visual-temporal context prompting},
  author={Cai, Shaofei and Wang, Zihao and Lian, Kewei and Mu, Zhancun and Ma, Xiaojian and Liu, Anji and Liang, Yitao},
  booktitle={Proceedings of the Computer Vision and Pattern Recognition Conference},
  pages={12122--12131},
  year={2025}
}

@inproceedings{ptgm,
  title={Pre-training goal-based models for sample-efficient reinforcement learning},
  author={Yuan, Haoqi and Mu, Zhancun and Xie, Feiyang and Lu, Zongqing},
  booktitle={The Twelfth International Conference on Learning Representations},
  year={2024}
}

@article{mimicplay,
  title={Mimicplay: Long-horizon imitation learning by watching human play},
  author={Wang, Chen and Fan, Linxi and Sun, Jiankai and Zhang, Ruohan and Fei-Fei, Li and Xu, Danfei and Zhu, Yuke and Anandkumar, Anima},
  journal={arXiv preprint arXiv:2302.12422},
  year={2023}
}

@article{gametars,
  title={Game-tars: Pretrained foundation models for scalable generalist multimodal game agents},
  author={Wang, Zihao and Li, Xujing and Ye, Yining and Fang, Junjie and Wang, Haoming and Liu, Longxiang and Liang, Shihao and Lu, Junting and Wu, Zhiyong and Feng, Jiazhan and others},
  journal={arXiv preprint arXiv:2510.23691},
  year={2025}
}

@InProceedings{pappas2025navigation,
  title = 	 {End-to-End Navigation with Vision-Language Models: Transforming Spatial Reasoning into Question-Answering},
  author =       {Goetting, Dylan and Singh, Himanshu Gaurav and Loquercio, Antonio},
  booktitle = 	 {Proceedings of the International Conference on Neuro-symbolic Systems},
  pages = 	 {22--35},
  year = 	 {2025},
  editor = 	 {Pappas, George and Ravikumar, Pradeep and Seshia, Sanjit A.},
  volume = 	 {288},
  series = 	 {Proceedings of Machine Learning Research},
  month = 	 {28--30 May},
  publisher =    {PMLR},
  url = 	 {https://proceedings.mlr.press/v288/goetting25a.html},
}

@article{rocket2,
  title={ROCKET-2: Steering Visuomotor Policy via Cross-View Goal Alignment},
  author={Cai, Shaofei and Mu, Zhancun and Liu, Anji and Liang, Yitao},
  journal={arXiv preprint arXiv:2503.02505},
  year={2025}
}

@article{dexgraspvla,
  title={Dexgraspvla: A vision-language-action framework towards general dexterous grasping},
  author={Zhong, Yifan and Huang, Xuchuan and Li, Ruochong and Zhang, Ceyao and Chen, Zhang and Guan, Tianrui and Zeng, Fanlian and Lui, Ka Num and Ye, Yuyao and Liang, Yitao and others},
  journal={arXiv preprint arXiv:2502.20900},
  year={2025}
}

@article{rdt,
  title={Rdt-1b: a diffusion foundation model for bimanual manipulation},
  author={Liu, Songming and Wu, Lingxuan and Li, Bangguo and Tan, Hengkai and Chen, Huayu and Wang, Zhengyi and Xu, Ke and Su, Hang and Zhu, Jun},
  journal={arXiv preprint arXiv:2410.07864},
  year={2024}
}

@article{octo,
  title={Octo: An open-source generalist robot policy},
  author={Team, Octo Model and Ghosh, Dibya and Walke, Homer and Pertsch, Karl and Black, Kevin and Mees, Oier and Dasari, Sudeep and Hejna, Joey and Kreiman, Tobias and Xu, Charles and others},
  journal={arXiv preprint arXiv:2405.12213},
  year={2024}
}

@article{rocket3,
  title={Scalable Multi-Task Reinforcement Learning for Generalizable Spatial Intelligence in Visuomotor Agents},
  author={Cai, Shaofei and Mu, Zhancun and Xia, Haiwen and Zhang, Bowei and Liu, Anji and Liang, Yitao},
  journal={arXiv preprint arXiv:2507.23698},
  year={2025}
}

@article{minestudio,
  title={MineStudio: A Streamlined Package for Minecraft AI Agent Development},
  author={Cai, Shaofei and Mu, Zhancun and He, Kaichen and Zhang, Bowei and Zheng, Xinyue and Liu, Anji and Liang, Yitao},
  journal={arXiv preprint arXiv:2412.18293},
  year={2024}
}

@inproceedings{grape,
  title={GRAPE: Generalizing Robot Policy via Preference Alignment},
  author={Zhang, Zijian and Zheng, Kaiyuan and Chen, Zhaorun and Jang, Joel and Li, Yi and Han, Siwei and Wang, Chaoqi and Ding, Mingyu and Fox, Dieter and Yao, Huaxiu},
  booktitle={ICRA 2025 Workshop on Foundation Models and Neuro-Symbolic AI for Robotics},
  year={2025}
}

@inproceedings{prismatic,
  title={Prismatic vlms: Investigating the design space of visually-conditioned language models},
  author={Karamcheti, Siddharth and Nair, Suraj and Balakrishna, Ashwin and Liang, Percy and Kollar, Thomas and Sadigh, Dorsa},
  booktitle={Forty-first International Conference on Machine Learning},
  year={2024}
}

@inproceedings{apo,
  title={Human-assisted Robotic Policy Refinement via Action Preference Optimization},
  author={Xia, Wenke and Yang, Yichu and Wu, Hongtao and Ma, Xiao and Kong, Tao and Hu, Di},
  booktitle={The Thirty-ninth Annual Conference on Neural Information Processing Systems},
  year={2025}
}

@article{vlarl,
  title={Vla-rl: Towards masterful and general robotic manipulation with scalable reinforcement learning},
  author={Lu, Guanxing and Guo, Wenkai and Zhang, Chubin and Zhou, Yuheng and Jiang, Haonan and Gao, Zifeng and Tang, Yansong and Wang, Ziwei},
  journal={arXiv preprint arXiv:2505.18719},
  year={2025}
}

@article{libero,
  title={Libero: Benchmarking knowledge transfer for lifelong robot learning},
  author={Liu, Bo and Zhu, Yifeng and Gao, Chongkai and Feng, Yihao and Liu, Qiang and Zhu, Yuke and Stone, Peter},
  journal={Advances in Neural Information Processing Systems},
  volume={36},
  pages={44776--44791},
  year={2023}
}

@inproceedings{softprompt,
  title={The Power of Scale for Parameter-Efficient Prompt Tuning},
  author={Lester, Brian and Al-Rfou, Rami and Constant, Noah},
  booktitle={Proceedings of the 2021 Conference on Empirical Methods in Natural Language Processing},
  pages={3045--3059},
  year={2021}
}

\newpage
\onecolumn
\appendix

\section{Theoretical Analysis: \alg\ as Frozen-Family (Manifold) Latent Control}
\label{sec:math}

We derive the objective optimized by \alg\ as a form of \emph{latent control} inside a \emph{frozen} goal-conditioned policy.
The key idea is to keep the policy parameters fixed and search over a low-dimensional conditioning variable~$g$
so as to match an energy-defined target behavior implied by preferences.

\subsection{Setup: Frozen goal-conditioned policy induces trajectory distributions}
\label{sec:setup}

Consider a controlled Markov process with state space $\mathcal{S}$, action space $\mathcal{A}$, horizon $T\in\mathbb{N}$,
initial-state distribution $p_0(s)$, and dynamics $p(s' \mid s,a)$.
Let $\pi_\theta(a\mid s,g)$ be a \emph{frozen} goal-conditioned policy with parameters $\theta$ and latent conditioning $g\in\mathcal{G}\subset\mathbb{R}^d$.

We will work with \emph{trajectory distributions} induced by $\pi_\theta(\cdot\mid\cdot,g)$.
For a length-$T$ trajectory
\[
\tau := (s_0,a_0,s_1,a_1,\ldots,s_{T-1},a_{T-1},s_T),
\]
define the induced trajectory density
\begin{equation}
p_{\theta,g}(\tau)
:= p_0(s_0)\prod_{t=0}^{T-1}\pi_\theta(a_t\mid s_t,g)\,p(s_{t+1}\mid s_t,a_t).
\label{eq:traj_dist_def}
\end{equation}
We also fix a \emph{reference} latent goal $g_{\mathrm{ref}}\in\mathcal{G}$ and define
\begin{equation}
p_{\mathrm{ref}}(\tau) := p_{\theta,g_{\mathrm{ref}}}(\tau).
\label{eq:pref_def}
\end{equation}

\paragraph{Assumption 1 (Shared environment dynamics).}
The conditioning $g$ affects actions only through the policy $\pi_\theta(a\mid s,g)$.
In particular, $p_0(s)$ and $p(s'\mid s,a)$ do not depend on $g$.
This ensures that likelihood ratios $p_{\theta,g}(\tau)/p_{\theta,g_{\mathrm{ref}}}(\tau)$ cancel all dynamics terms.

\subsection{Frozen behavioral family (``manifold'') and safety prior}
\label{sec:frozen_family}

The frozen policy backbone induces a family of behaviors indexed by $g$:
\begin{equation}
\Pi_\theta := \{\pi_\theta(\cdot\mid\cdot,g)\,:\, g\in\mathcal{G}\}.
\label{eq:family_def}
\end{equation}
(One may call $\Pi_\theta$ a ``behavioral manifold'' only after specifying a topology/geometry under which the map
$g\mapsto \pi_\theta(\cdot\mid\cdot,g)$ is a smooth embedding; we will only require it as a restricted family.)

\paragraph{Assumption 2 (Support/safety prior).}
For all $g\in\mathcal{G}$, trajectories sampled from $p_{\theta,g}(\tau)$ remain in the support of valid,
physically coherent behaviors captured by pretraining.
Thus adaptation is posed as \emph{search over} $g$ rather than changing $\theta$.

\subsection{Energy-based target distribution relative to a reference prior}
\label{sec:ebm}

We encode the task not via an explicit reward but via an (unknown) energy functional
$E_{\mathrm{task}}(\tau)\in\mathbb{R}$, observed only through preferences.
Following the standard maximum-entropy / control-as-inference form, define the \emph{target} trajectory distribution
\begin{equation}
p^\star(\tau)
:= \frac{1}{Z(\beta)}\,p_{\mathrm{ref}}(\tau)\,\exp\!\bigl(-\beta\,E_{\mathrm{task}}(\tau)\bigr),
\qquad \beta>0,
\label{eq:boltzmann}
\end{equation}
where $Z(\beta)$ is the partition function.
The temperature $\beta$ sets the energy scale; in preference modeling it also plays the role of a noise/inconsistency scale,
so it can be treated as a hyperparameter (or estimated) without loss of generality.

\subsection{Latent control as (approximate) reweighting: realizability or projection}
\label{sec:realizability}

The family $\{p_{\theta,g}\}_{g\in\mathcal{G}}$ generally cannot represent $p^\star$ exactly.
We therefore separate two cases:

\paragraph{(i) Realizability (strong form).}
There exists $g^\star\in\mathcal{G}$ such that $p_{\theta,g^\star}(\tau)=p^\star(\tau)$ for all $\tau$.
Under realizability, Eq.~\eqref{eq:boltzmann} implies
\begin{equation}
-\beta E_{\mathrm{task}}(\tau)
= \log\frac{p_{\theta,g^\star}(\tau)}{p_{\mathrm{ref}}(\tau)} + \log Z(\beta).
\label{eq:energy_reparam_realizable}
\end{equation}

\paragraph{(ii) Projection (weak form; what we optimize).}
Without realizability, we interpret \alg\ as selecting
\begin{equation}
g^\star \in \arg\min_{g\in\mathcal{G}} D_{\mathrm{KL}}\!\bigl(p^\star(\tau)\,\|\,p_{\theta,g}(\tau)\bigr),
\label{eq:kl_projection}
\end{equation}
or equivalently as maximum-likelihood fitting of a preference model (next subsection).
In either case, the natural \emph{model} energy induced by $g$ is the log-likelihood ratio (up to an additive constant):
\begin{equation}
\widehat{E}_g(\tau)
:= -\frac{1}{\beta}\log\frac{p_{\theta,g}(\tau)}{p_{\mathrm{ref}}(\tau)}
\quad\text{(defined up to an additive constant in $\tau$)}.
\label{eq:model_energy}
\end{equation}
This definition makes explicit the ``latent control'' interpretation: changing $g$ reweights the reference prior
by a trajectory-wise likelihood ratio.

\subsection{Preferences via Bradley--Terry and the DPO-like objective over \texorpdfstring{$g$}{g}}
\label{sec:energy_to_preference}

We observe only pairwise preferences $\mathcal{D}=\{(\tau_w,\tau_l)\}$, where $\tau_w$ is preferred to $\tau_l$.
Under the Bradley--Terry--Luce (BTL) / logistic choice model,
\begin{equation}
\mathbb{P}(\tau_w \succ \tau_l)
= \sigma\!\Bigl(-\beta\bigl(E_{\mathrm{task}}(\tau_w)-E_{\mathrm{task}}(\tau_l)\bigr)\Bigr),
\qquad \sigma(u)=\frac{1}{1+e^{-u}}.
\label{eq:btl}
\end{equation}

If realizability holds, substitute Eq.~\eqref{eq:energy_reparam_realizable} and note that $\log Z(\beta)$ cancels between $\tau_w,\tau_l$.
More generally, in the projection view we \emph{model} energies via $\widehat{E}_g$ from Eq.~\eqref{eq:model_energy},
leading to the model preference probability
\begin{align}
\mathbb{P}_g(\tau_w \succ \tau_l)
&= \sigma\!\Bigl(-\beta(\widehat{E}_g(\tau_w)-\widehat{E}_g(\tau_l))\Bigr) \nonumber\\
&= \sigma\!\left(
\log\frac{p_{\theta,g}(\tau_w)}{p_{\mathrm{ref}}(\tau_w)}
-
\log\frac{p_{\theta,g}(\tau_l)}{p_{\mathrm{ref}}(\tau_l)}
\right).
\label{eq:model_pref_prob}
\end{align}
Maximizing likelihood over $\mathcal{D}$ (equivalently minimizing negative log-likelihood) yields the trajectory-wise DPO-like loss
\begin{equation}
\mathcal{L}(g)
:= -\mathbb{E}_{(\tau_w,\tau_l)\sim\mathcal{D}}
\left[
\log \sigma\!\left(
\log\frac{p_{\theta,g}(\tau_w)}{p_{\mathrm{ref}}(\tau_w)}
-
\log\frac{p_{\theta,g}(\tau_l)}{p_{\mathrm{ref}}(\tau_l)}
\right)
\right].
\label{eq:traj_dpo}
\end{equation}

\subsection{Markov factorization and step-wise log-ratio form}
\label{sec:markov_decomp}

Using Eq.~\eqref{eq:traj_dist_def} and Assumption~1 (shared dynamics), the trajectory likelihood ratio becomes
\begin{equation}
\log\frac{p_{\theta,g}(\tau)}{p_{\mathrm{ref}}(\tau)}
=
\sum_{t=0}^{T-1}\log\frac{\pi_\theta(a_t\mid s_t,g)}{\pi_\theta(a_t\mid s_t,g_{\mathrm{ref}})},
\label{eq:ratio_decomp}
\end{equation}
since $p_0$ and $p(\cdot\mid\cdot,\cdot)$ cancel identically.
Substituting Eq.~\eqref{eq:ratio_decomp} into Eq.~\eqref{eq:traj_dpo} yields the practical step-wise objective
\begin{equation}
\mathcal{L}(g)
= -\mathbb{E}_{(\tau_w,\tau_l)\sim\mathcal{D}}
\left[
\log \sigma\!\left(
\sum_{t=0}^{T-1}
\left[
\log\frac{\pi_\theta(a_t^{(w)}\mid s_t^{(w)},g)}{\pi_\theta(a_t^{(w)}\mid s_t^{(w)},g_{\mathrm{ref}})}
-
\log\frac{\pi_\theta(a_t^{(l)}\mid s_t^{(l)},g)}{\pi_\theta(a_t^{(l)}\mid s_t^{(l)},g_{\mathrm{ref}})}
\right]
\right)
\right].
\label{eq:final_pgt_loss}
\end{equation}

\subsection{Reduction to generic pairwise preference learning}
\label{sec:math_pref_framework}

To connect with standard analyses of preference learning, abstract a generic \emph{context} $x$ and outputs $(y^+,y^-)$.
Let a model $p_\varphi(y\mid x)$ and reference $p_{\mathrm{ref}}(y\mid x)$ be given, and define
\begin{align}
\Delta_\varphi(x;y^+,y^-) &:= \log p_\varphi(y^+\mid x)-\log p_\varphi(y^-\mid x),\\
\Delta_{\mathrm{ref}}(x;y^+,y^-) &:= \log p_{\mathrm{ref}}(y^+\mid x)-\log p_{\mathrm{ref}}(y^-\mid x),\\
h_\varphi(x;y^+,y^-) &:= \Delta_\varphi(x;y^+,y^-)-\Delta_{\mathrm{ref}}(x;y^+,y^-)
= \log\frac{p_\varphi(y^+\mid x)\,p_{\mathrm{ref}}(y^-\mid x)}{p_\varphi(y^-\mid x)\,p_{\mathrm{ref}}(y^+\mid x)}.
\label{eq:h_def}
\end{align}
The per-pair DPO loss is
\begin{equation}
\ell_{\mathrm{DPO}}(h;\beta):=-\log\sigma(\beta h)=\log(1+e^{-\beta h}),\qquad \beta>0.
\label{eq:dpo_perpair}
\end{equation}
Our trajectory objective in Eq.~\eqref{eq:traj_dpo} is exactly this form with the instantiation
\[
\varphi \leftarrow g,\quad y\leftarrow \tau,\quad x\leftarrow \text{(initial state / prompt / context)},\quad
p_\varphi(\cdot\mid x)\leftarrow p_{\theta,g}(\cdot\mid x),\quad p_{\mathrm{ref}}\leftarrow p_{\theta,g_{\mathrm{ref}}}.
\]

\subsubsection{\texorpdfstring{DPO $\Rightarrow$ IPO}{DPO => IPO} via a second-order expansion with a uniform remainder bound}
\label{sec:dpo2ipo}

Let $f(h):=\log(1+e^{-\beta h})$. Then
\begin{align}
f'(h) &= -\beta\,\sigma(-\beta h),\\
f''(h) &= \beta^2\,\sigma(\beta h)\bigl(1-\sigma(\beta h)\bigr),\\
f^{(3)}(h) &= \beta^3\,\sigma(\beta h)\bigl(1-\sigma(\beta h)\bigr)\bigl(1-2\sigma(\beta h)\bigr).
\end{align}
Since $\sigma(t)(1-\sigma(t))\le 1/4$ and $|1-2\sigma(t)|\le 1$, we have the global bound
\begin{equation}
\sup_{h\in\mathbb{R}} |f^{(3)}(h)| \le \frac{\beta^3}{4}.
\label{eq:third_deriv_bound}
\end{equation}

\begin{theorem}[Second-order Taylor approximation of DPO with a uniform cubic remainder bound]
\label{theorem:second_order_taylor}
For any $h\in\mathbb{R}$,
\begin{equation}
\ell_{\mathrm{DPO}}(h;\beta)
= \log 2 - \frac{\beta}{2}h + \frac{\beta^2}{8}h^2 + R_3(h),
\qquad
|R_3(h)| \le \frac{\beta^3}{24}|h|^3.
\label{eq:dpo_taylor_bound}
\end{equation}
Equivalently, completing the square yields
\begin{equation}
\ell_{\mathrm{DPO}}(h;\beta)
= \frac{\beta^2}{8}\Bigl(h-\frac{2}{\beta}\Bigr)^2 + \Bigl(\log 2-\frac{1}{2}\Bigr) + R_3(h),
\qquad
|R_3(h)| \le \frac{\beta^3}{24}|h|^3.
\label{eq:dpo_square_bound}
\end{equation}
\end{theorem}

\begin{proof}
Apply Taylor's theorem with Lagrange remainder at $0$:
$f(h)=f(0)+f'(0)h+\tfrac12 f''(0)h^2+\tfrac16 f^{(3)}(\xi)h^3$ for some $\xi$ between $0$ and $h$.
Compute $f(0)=\log2$, $f'(0)=-\beta/2$, $f''(0)=\beta^2/4$, and use \eqref{eq:third_deriv_bound}.
\end{proof}

The sampled IPO objective uses a squared loss
\begin{equation}
\ell_{\mathrm{IPO}}(h;\tau):=\tau\Bigl(h-\frac{1}{2\tau}\Bigr)^2,\qquad \tau>0.
\label{eq:ipo_sampled}
\end{equation}
Setting $\tau=\beta/4$ aligns the minimizer of the quadratic approximation (at $h^\star=2/\beta$) with IPO's target $h^\star=1/(2\tau)$, and yields
\begin{equation}
\frac{\beta^2}{8}\Bigl(h-\frac{2}{\beta}\Bigr)^2
= \frac{\beta}{2}\,\ell_{\mathrm{IPO}}(h;\beta/4).
\label{eq:dpo_ipo_scaling}
\end{equation}

\subsubsection{\texorpdfstring{DPO $\Rightarrow$ hinge}{DPO => hinge} ranking term via sandwich bounds and a hard-margin limit}
\label{sec:dpo2hinge}

Define $H(z):=\max(0,-z)$. For all $t\in\mathbb{R}$,
\begin{equation}
\max(0,-t)\ \le\ \log(1+e^{-t})\ \le\ \max(0,-t)+\log 2.
\label{eq:softplus_hinge_sandwich}
\end{equation}
Applying \eqref{eq:softplus_hinge_sandwich} to $t=\beta z$ gives
\begin{equation}
0\ \le\ \ell_{\mathrm{DPO}}(z;\beta) - \beta H(z)\ \le\ \log 2,
\qquad \forall z\in\mathbb{R},
\label{eq:dpo_hinge_uniform_bound}
\end{equation}
and hence
\begin{equation}
0\ \le\ \frac{1}{\beta}\ell_{\mathrm{DPO}}(z;\beta) - H(z)\ \le\ \frac{\log 2}{\beta}.
\label{eq:dpo_hinge_normalized}
\end{equation}
Thus for fixed $z$, $\lim_{\beta\to\infty}\frac{1}{\beta}\ell_{\mathrm{DPO}}(z;\beta)=H(z)$.

Since $h_\varphi=\Delta_\varphi-\Delta_{\mathrm{ref}}$, the hard-margin limit yields the hinge ranking term
\begin{equation}
H(h_\varphi)=\max(0,-h_\varphi)=\max\bigl(0,\ \Delta_{\mathrm{ref}}-\Delta_\varphi\bigr).
\label{eq:hinge_with_ref_margin}
\end{equation}
If one replaces the (possibly sample-dependent) margin $\Delta_{\mathrm{ref}}$ by a chosen constant $\delta$, this becomes
\begin{equation}
\ell_{\mathrm{rank}}(\varphi)
=\max\bigl(0,\ \delta - \log p_\varphi(y^+\mid x) + \log p_\varphi(y^-\mid x)\bigr),
\label{eq:slic_rank}
\end{equation}
which matches the SLiC-HF ranking/calibration term.

\subsubsection{SLiC-HF regularizer as a Lagrangian relaxation with an exact reverse-KL interpretation}
\label{sec:slic_hf_regularizer}

SLiC-HF adds a cross-entropy term on a target $y_{\mathrm{ref}}$:
\begin{equation}
\ell_{\mathrm{SLiC\mbox{-}HF}}(\varphi)
=\max\bigl(0,\delta-\log p_\varphi(y^+\mid x)+\log p_\varphi(y^-\mid x)\bigr)
\;-\;\lambda \log p_\varphi(y_{\mathrm{ref}}\mid x).
\label{eq:slic_full}
\end{equation}

\begin{proposition}[Cross-entropy regularization equals reverse KL up to a constant]
\label{prop:ce_is_reverse_kl}
Let $x\sim\rho(x)$ and $y\sim p_{\mathrm{ref}}(y\mid x)$. Then
\begin{equation}
\mathbb{E}\bigl[-\log p_\varphi(y\mid x)\bigr]
=
D_{\mathrm{KL}}\!\bigl(p_{\mathrm{ref}}(\cdot\mid x)\,\|\,p_\varphi(\cdot\mid x)\bigr)
+\mathbb{H}\!\bigl(p_{\mathrm{ref}}(\cdot\mid x)\bigr),
\label{eq:ce_reverse_kl}
\end{equation}
where $\mathbb{H}(p_{\mathrm{ref}}(\cdot\mid x))$ is independent of $\varphi$.
\end{proposition}

\begin{proof}
By definition,
$D_{\mathrm{KL}}(p_{\mathrm{ref}}\|p_\varphi)
=\mathbb{E}_{p_{\mathrm{ref}}}[\log p_{\mathrm{ref}}-\log p_\varphi]
=\mathbb{E}_{p_{\mathrm{ref}}}[-\log p_\varphi]-\mathbb{E}_{p_{\mathrm{ref}}}[-\log p_{\mathrm{ref}}]$.
Rearrange and note $\mathbb{E}_{p_{\mathrm{ref}}}[-\log p_{\mathrm{ref}}]=\mathbb{H}(p_{\mathrm{ref}})$.
\end{proof}

\paragraph{Exact Lagrangian form.}
Consider the constrained problem
\begin{equation}
\min_\varphi\ \mathbb{E}\bigl[\ell_{\mathrm{rank}}(\varphi)\bigr]
\quad\text{s.t.}\quad
\mathbb{E}_{(x,y)\sim \rho\,p_{\mathrm{ref}}}\bigl[-\log p_\varphi(y\mid x)\bigr]\ \le\ c.
\label{eq:constrained_rank_ce}
\end{equation}
Its Lagrangian relaxation is
\begin{equation}
\min_\varphi\ \mathbb{E}\bigl[\ell_{\mathrm{rank}}(\varphi)\bigr]
+\lambda\Bigl(\mathbb{E}_{\rho\,p_{\mathrm{ref}}}[-\log p_\varphi(y\mid x)]-c\Bigr),
\qquad \lambda\ge 0,
\label{eq:lagrangian_rank_ce}
\end{equation}
which matches \eqref{eq:slic_full} up to constants. By Proposition~\ref{prop:ce_is_reverse_kl}, the regularizer equals
a reverse-KL penalty $D_{\mathrm{KL}}(p_{\mathrm{ref}}\|p_\varphi)$ plus an additive constant, precisely formalizing
``staying close to the reference.''

\paragraph{Implications for \alg.}
Instantiating $\varphi\leftarrow g$ and $p_\varphi\leftarrow p_{\theta,g}$ shows that DPO/IPO/SLiC-HF are interchangeable
preference-fitting modules for selecting $g$ inside a frozen family.
They share the same core logit $h$ in Eq.~\eqref{eq:h_def}, differing mainly in (i) the convex surrogate (logistic/quadratic/hinge)
and (ii) whether reference-closeness is enforced implicitly via $\Delta_{\mathrm{ref}}$ or explicitly via a KL-equivalent regularizer.

\section{Experiment Details}
\label{expriment}
\subsection{Minecraft}
Minecraft is a popular sandbox game that allows players to freely create and explore their world. Since Minecraft is an open-world environment, many recent works have designed agents and conducted explorations within Minecraft \citep{malmo}. In this work, we conduct experiments on 1.16.5 version MineRL \citep{minerl} and MCP-Reborn. 

\subsection{Minecraft SkillForge Benchmark}
\label{skillforge}
Minecraft SkillForge Benchmark is a comprehensive task suite that covers various types of tasks in Minecraft. All tasks are categorized into six major groups:
\begin{itemize}[nosep]
    \item Collect task: these tasks are designed to evaluate an AI agent’s capability in resource acquisition proficiency and spatial awareness. 
    \item Craft task: these tasks are designed to shed light on an AI agent’s prowess in item utilization, the intricacies of Minecraft crafting mechanics, and the nuances of various game mechanic interactions. 
    \item Explore task: these tasks are designed to evaluate an AI agent’s navigation proficiency, understanding of diverse environments, and intrinsic motivation for exploration. 
    \item Survive task: these tasks are designed to analyze an AI agent’s ability to ensure its survival, adeptness in combat scenarios, and capability to interact with the environment to meet basic needs. 
    \item Tool task: these tasks are designed to deeply investigate an AI agent’s capabilities in tool utilization, precision in tool handling, and contextual application of various tools to carry out specific tasks. 
    \item Build task: these tasks are devised to evaluate an AI agent’s aptitude in structural reasoning, spatial organization, and its capability to interact with and manipulate the environment to create specific structures or outcomes.
\end{itemize}

\subsection{Task Metrics and Selection}
\label{taskselection}
For most tasks, the environment logs the rewards when the corresponding objectives are achieved. We define tasks with a reward function greater than 0 as successful, and the frequency of successfully completing a task is referred to as the success rate. However, tasks like ``collect\_wood'' ``explore\_mine'' and ``survive\_plant'' have a success rate of over 95\% across different agents, and the specific values of the reward function are meaningful, reflecting the agents' capabilities in these tasks, so we use the detailed reward value as the metric.

We removed the tasks that are too easy that agents can perform a success rate of 100\% while the specific value of the reward is either high enough (e.g. collect\_grass) or not meaningful (e.g. survive\_sleep). Also, to simplify the experiment, We removed the tasks for which the reward function cannot be directly obtained from the game, including subjective tasks (e.g. building tasks) and objective tasks where the environment does not log explicit rewards (e.g. craft\_smelt). Moreover, mining obsidian is a high requirement for the agent's sensitivity to the objectives, and the agent needs to stay focused on the same goal over extended time steps to perform useful actions; therefore, we consider this task to be quite important and add it to the testing tasks apart from \emph{Minecraft SkillForge}.

\subsection{Out-of-distribution Settings}
\label{OOD}
We designed the out-of-distribution (OOD) setting with the goal of preventing the policy from overfitting to the environment and relying on it to dictate behavior. Thus, without altering the core meaning of the tasks, we made the following modifications to create the OOD setting:
\begin{itemize}[nosep]
    \item \textbf{Seed and agent location} We change the seed and spawn location in the Minecraft world to perform the same task, and then the initial observation will not be identical to the training set.
    \item \textbf{Biome} We change the biome of the agent while keeping the task solvable. For example, change biome from plains to forest of task \texttt{tool\_pumpkin}(\includegraphics[scale=0.043,valign=c]{sections/assets/minecraft/pumpkin.png}).
    \item \textbf{Tool} We modified the auxiliary tools while ensuring the tasks remained solvable. For example, in the \texttt{survive\_hunt}(\includegraphics[scale=0.05,valign=c]{sections/assets/minecraft/hunt.jpg}), we replaced the iron\_sword with diamond\_axe.
    \item \textbf{Object location} We change the location of the object that the agent needs to interact with. For example, we changed the position of the stonecutter from being held in the hand to being placed in front of the agent.
\end{itemize}
For each task, we applied one or more of the aforementioned OOD modifications. It is important to note that the absolute performance in the OOD setting is not directly comparable to the baseline, as the tasks may become either easier or harder in the OOD environment.

\subsection{Hyperparameters}
\label{sec:hp}
Our training hyperparameters in Minecraft settings are listed in Table \ref{tab:hp_minecraft}. Each in-distribution experiment is trained and evaluated across three distinct scenarios, whereas out-of-distribution experiments are trained on the same scenarios but evaluated on a different set of three unseen scenarios. The reported test performance of GROOT and STEVE is averaged over 1,000 evaluation runs, while the results for GROOT+ and STEVE+ are averaged over 500 runs. In certain tasks, scenario generation involves manual intervention rather than relying solely on constraints natively supported by the game simulator; for example, in the \texttt{survive\_combat}(\includegraphics[scale=0.055,valign=c]{sections/assets/minecraft/combat.jpg}) task, mobs are explicitly spawned instead of emerging naturally from the environment.

\begin{table*}[h]
\centering
\caption{Hyperparameters for training. } \label{tab:hp_minecraft}
\begin{tabular}{@{}cc@{}}
\toprule
\textbf{Hyperparameter}  & \textbf{Value}   \\ \midrule
Optimizer                & Adam     \\
Learning Rate            & 1e-2 \\
$\beta$ (in DPO)         & 0.6 \\
Batch Size               & Full Gradient Descent \\
Type of GPUs             & NVIDIA RTX 4090 \\
Training Precision       & float32      \\
Number of P-N Samples (each)       & 180 \\
\bottomrule
\end{tabular}
\end{table*}

Our training hyperparameters in OpenVLA LIBERO-goal settings are listed in Table \ref{tab:hp_openvla}. Exception: We observed that OpenVLA fails excessively on Task 3 of the LIBERO-goal benchmark. To improve training stability, we increased the value of $\beta$ to 0.5.

\begin{table*}[h]
\centering
\caption{Hyperparameters for training. } \label{tab:hp_openvla}
\begin{tabular}{@{}cc@{}}
\toprule
\textbf{Hyperparameter}  & \textbf{Value}   \\ \midrule
Optimizer                & Adam     \\
Learning Rate            & 1e-5 \\
$\beta$ (in DPO)         & 0.3 \\
Batch Size               & Full Gradient Descent \\
Type of GPUs             & NVIDIA A800 \\
Training Precision       & bfloat16      \\
Number of P-N Samples (each)       & 10 \\
\bottomrule
\end{tabular}
\end{table*}

\section{Experiment Results}

\subsection{Behaviour Cloning Results}
This baseline employs behavior cloning, trained exclusively on positive samples, without the inclusion of negative data or preference learning. We present results for both tuning the latent goal only and the full parameters (Table \ref{tab:BC}).

\subsection{Full Fine-tuning Results}
\label{fpft_detail}
We compare the results of our method with full fine-tuning. The latter involves $\sim$100M parameters, while the former only has 512 parameters, which is merely one in hundreds of thousands of the other. We found that in in-distribution settings, \alg achieves results comparable to those of full fine-tuning. However, in out-of-distribution (OOD) environments, \alg outperformed across all tasks. The result can be found in Table \ref{tab:fpft}.
\begin{table}[]
\centering
\caption{\textbf{Comparisons between full fine-tuning and \alg.} The ``soft-prompt-like''~\citep{softprompt, dualprompt} method can bring better improvements than the counterpart, especially on OOD settings.}
\label{tab:fpft}
\resizebox{0.55\linewidth}{!}{
\renewcommand\arraystretch{1.1}
\begin{tabular}{@{}cccclccc@{}}
\toprule
\multirow{2}{*}{Task} & \multicolumn{3}{c}{In Distribution (ID)} &  & \multicolumn{3}{c}{Out of Distribution (OOD)} \\ \cmidrule(lr){2-4} \cmidrule(l){6-8} 
 & Pretrained & Full & \alg &  & Pretrained & Full & \alg \\ \midrule
\includegraphics[scale=0.043,valign=c]{sections/assets/minecraft/log.png} & 3.14 & 3.46 & \textbf{3.62} &  & 3.88 & 4.04 & \textbf{4.22} \\
\includegraphics[scale=0.045,valign=c]{sections/assets/minecraft/stonecutter.png} & 31.0 & \textbf{62.2} & 44.6 &  & 20.0 & 21.2 & \textbf{23.4} \\
\includegraphics[scale=0.06,valign=c]{sections/assets/minecraft/stone.jpg} & 4.91 & 6.00 & \textbf{6.58} &  & 3.90 & 4.77 & \textbf{5.38} \\
\includegraphics[scale=0.05,valign=c]{sections/assets/minecraft/hunt.jpg} & 31.2 & \textbf{39.8} & \textbf{39.8} &  & 20.8 & 21.0 & \textbf{21.6} \\
\includegraphics[scale=0.043,valign=c]{sections/assets/minecraft/pumpkin.png} & 48.3 & \textbf{58.4} & 57.8 &  & 16.6 & 22.2 & \textbf{25.8} \\
\includegraphics[scale=0.7,valign=c]{sections/assets/minecraft/obsidian.png} & 42.0 & \textbf{62.2} & 57.2 &  & 4.2 & 6.0 & \textbf{8.2} \\ \bottomrule
\end{tabular}}
\end{table}

\subsection{Parameter-efficient Fine-tuning Results}
\label{pe}
We conduct parameter-efficient fine-tuning on LoRA \citep{lora}, BitFit \citep{bitfit}, VeRA \citep{vera}, and the result is in Table \ref{tab:pe}. In fact, all of these parameter counts are significantly larger than those of \alg. and the contrast is shown in Table \ref{tab:param_num}.
\begin{table}[]
\centering
\caption{Parameter-efficient fine-tuning result.}
\label{tab:pe}
\resizebox{0.65\linewidth}{!}{
\renewcommand\arraystretch{1.1}
\begin{tabular}{@{}cccccccccc@{}}
\toprule
\multirow{2}{*}{Task} & \multicolumn{4}{c}{In Distribution(ID)} &  & \multicolumn{4}{c}{Out of Distribution(OOD)} \\ \cmidrule(lr){2-5} \cmidrule(l){7-10} 
 & LoRA & BitFit & VeRA & \alg &  & LoRA & BitFit & VeRA & \alg \\ \midrule
\includegraphics[scale=0.043,valign=c]{sections/assets/minecraft/log.png} & 3.47 & 3.55 & 3.39 & \textbf{3.62} &  & 4.09 & 3.91 & 4.16 & \textbf{4.22} \\
\includegraphics[scale=0.045,valign=c]{sections/assets/minecraft/stonecutter.png} & 49.4 & 48.6 & \textbf{52.2} & 44.6 &  & 19.8 & 18.0 & 18.8 & \textbf{23.4} \\
\includegraphics[scale=0.06,valign=c]{sections/assets/minecraft/stone.jpg} & 6.52 & 5.37 & 5.76 & \textbf{6.58} &  & 5.17 & 4.42 & 4.67 & \textbf{5.38} \\
\includegraphics[scale=0.05,valign=c]{sections/assets/minecraft/hunt.jpg} & 39.8 & 40.8 & \textbf{42.0} & 39.8 &  & 24.6 & 25.2 & \textbf{27.4} & 21.6 \\
\includegraphics[scale=0.043,valign=c]{sections/assets/minecraft/pumpkin.png} & 50.4 & 56.2 & 52.8 & \textbf{57.8} &  & 19.6 & 20.8 & 22.4 & \textbf{25.8} \\
\includegraphics[scale=0.7,valign=c]{sections/assets/minecraft/obsidian.png} & \textbf{71.2} & 55.8 & 57.8 & 57.2 &  & \textbf{10.6} & 6.2 & 2.6 & 8.2 \\ \bottomrule
\end{tabular}}
\end{table}
\begin{table}[h]
\centering
\caption{The number of trainable parameters in full fine-tuning, \alg and other baselines.}
\resizebox{0.5\linewidth}{!}{
\renewcommand\arraystretch{1.2}
\begin{tabular}{cccccc}
\toprule
& Full & LoRA & BitFit & VeRA & \alg \\ \hline
\textbf{\# Parameters} & 86M & 393K & 80K & 15K & 512 \\
\bottomrule
\end{tabular}}
\label{tab:param_num}
\end{table}

\subsection{Continual Learning Results}
\label{CL}
All of our continual learning baselines are based on fine-tuning the entire policy model, and the order of tasks for continual learning is as follows:  \texttt{collect\_obsidian}(\includegraphics[scale=0.6,valign=c]{sections/assets/minecraft/obsidian.png}) $\to$ \texttt{tool pumpkin}(\includegraphics[scale=0.04,valign=c]{sections/assets/minecraft/carved_pumpkin.png}) $\to$ \texttt{craft\_crafting\_table}(\includegraphics[scale=0.05,valign=c]{sections/assets/minecraft/table.jpeg}) $\to$ \texttt{explore climb}(\includegraphics[scale=0.035,valign=c]{sections/assets/minecraft/Steve.png}). We implemented multi-task learning (MTL) (Table \ref{tab:MTL}), naive continual learning (NCL) (Table \ref{tab:NCL}), knowledge distillation (KD)(Table \ref{tab:KD}), experience replay (ER)(Table \ref{tab:ER}), and elastic weight consolidation (EWC)(Table \ref{tab:EWC}).
\begin{table}[h]
\centering
\caption{\textbf{Task Sequential Adaptation: Continual Learning with Naive Continual Learning.} The task names in the first row represent the model trained up to the current task during sequential training (with both the pretrained model and \alg used as references); the task names in the first column represent the test results on each task. To reduce human annotation costs, we do not test the results of $\texttt{explore\_climb}$, but use it solely as a step in the training process and reduce the number of trajectory pairs to 40. It is employed to examine the impact of later tasks on earlier ones during sequential training. The same principle applies to the subsequent tables on continual learning.}
\resizebox{0.5\linewidth}{!}{
\renewcommand\arraystretch{1.2}
\begin{tabular}{ccccccc}
\toprule
Task & \includegraphics[scale=0.65,valign=c]{sections/assets/minecraft/obsidian.png} & \includegraphics[scale=0.043,valign=c]{sections/assets/minecraft/pumpkin.png} & \includegraphics[scale=0.055,valign=c]{sections/assets/minecraft/table.jpeg} & \includegraphics[scale=0.05,valign=c]{sections/assets/minecraft/Steve.png} & Pretrained & \alg(Ours) \\ \hline
\includegraphics[scale=0.7,valign=c]{sections/assets/minecraft/obsidian.png} & 6.0 & 4.6 & 7.0 & 6.8 & 4.2 & \textbf{8.2} \\
\includegraphics[scale=0.043,valign=c]{sections/assets/minecraft/pumpkin.png} &  & 23.6 & 24.2 & 20.4 & 16.6 & \textbf{25.8} \\
\includegraphics[scale=0.055,valign=c]{sections/assets/minecraft/table.jpeg} &  &  & 5.2 & 7.0 & 6.0 & \textbf{18.4} \\
\bottomrule
\end{tabular}}
\label{tab:NCL}
\end{table}
\begin{table}[h]
\centering
\caption{Task Sequential Adaptation: Continual Learning with Elastic Knowledge Distillation}
\resizebox{0.5\linewidth}{!}{
\renewcommand\arraystretch{1.2}
\begin{tabular}{ccccccc}
\toprule
Task & \includegraphics[scale=0.7,valign=c]{sections/assets/minecraft/obsidian.png} & \includegraphics[scale=0.043,valign=c]{sections/assets/minecraft/pumpkin.png} & \includegraphics[scale=0.055,valign=c]{sections/assets/minecraft/table.jpeg} & \includegraphics[scale=0.05,valign=c]{sections/assets/minecraft/Steve.png} & Pretrained & \alg(Ours) \\ \hline
\includegraphics[scale=0.7,valign=c]{sections/assets/minecraft/obsidian.png} & 6.0 & 5.2 & 6.6 & 5.4 & 4.2 & \textbf{8.2} \\
\includegraphics[scale=0.043,valign=c]{sections/assets/minecraft/pumpkin.png} &  & 24.6 & 23.4 & 20.6 & 16.6 & \textbf{25.8} \\
\includegraphics[scale=0.055,valign=c]{sections/assets/minecraft/table.jpeg} &  &  & 7.6 & 5.8 & 6.0 & \textbf{18.4} \\
\bottomrule
\end{tabular}}
\label{tab:KD}
\end{table}
\begin{table}[h]
\centering
\caption{Task Sequential Adaptation: Continual Learning with Experience Replay}
\resizebox{0.5\linewidth}{!}{
\renewcommand\arraystretch{1.2}
\begin{tabular}{ccccccc}
\toprule
Task & \includegraphics[scale=0.7,valign=c]{sections/assets/minecraft/obsidian.png} & \includegraphics[scale=0.043,valign=c]{sections/assets/minecraft/pumpkin.png} & \includegraphics[scale=0.055,valign=c]{sections/assets/minecraft/table.jpeg} & \includegraphics[scale=0.05,valign=c]{sections/assets/minecraft/Steve.png} & Pretrained & \alg(Ours) \\ \hline
\includegraphics[scale=0.7,valign=c]{sections/assets/minecraft/obsidian.png} & 6.0 & 6.6 & 5.0 & 6.0 & 4.2 & \textbf{8.2} \\
\includegraphics[scale=0.043,valign=c]{sections/assets/minecraft/pumpkin.png} &  & 22.8 & 21.8 & 25.0 & 16.6 & \textbf{25.8} \\
\includegraphics[scale=0.055, valign=c]{sections/assets/minecraft/table.jpeg} &  &  & 5.2 & 9.0 & 6.0 & \textbf{18.4} \\
\bottomrule
\end{tabular}}
\label{tab:ER}
\end{table}
\begin{table}[h]
\centering
\caption{Task Sequential Adaptation: Continual Learning with Elastic Weight Consolidation}
\resizebox{0.5\linewidth}{!}{
\renewcommand\arraystretch{1.2}
\begin{tabular}{ccccccc}
\toprule
Task & \includegraphics[scale=0.7,valign=c]{sections/assets/minecraft/obsidian.png} & \includegraphics[scale=0.043,valign=c]{sections/assets/minecraft/pumpkin.png} & \includegraphics[scale=0.055,valign=c]{sections/assets/minecraft/table.jpeg} & \includegraphics[scale=0.05,valign=c]{sections/assets/minecraft/Steve.png} & Pretrained & \alg(Ours) \\ \hline
\includegraphics[scale=0.7,valign=c]{sections/assets/minecraft/obsidian.png} & 6.0 & 8.2 & 5.4 & 5.4 & 4.2 & \textbf{8.2} \\
\includegraphics[scale=0.043,valign=c]{sections/assets/minecraft/pumpkin.png} &  & 23.6 & 24.0 & 23.8 & 16.6 & \textbf{25.8} \\
\includegraphics[scale=0.055,valign=c]{sections/assets/minecraft/table.jpeg} &  &  & 5.0 & 7.4 & 6.0 & \textbf{18.4} \\
\bottomrule
\end{tabular}}
\label{tab:EWC}
\end{table}

\subsection{The Selection of \texorpdfstring{$\beta$}{beta}}

\begin{figure*}[t]
    \centering
    \includegraphics[width=0.5\linewidth]{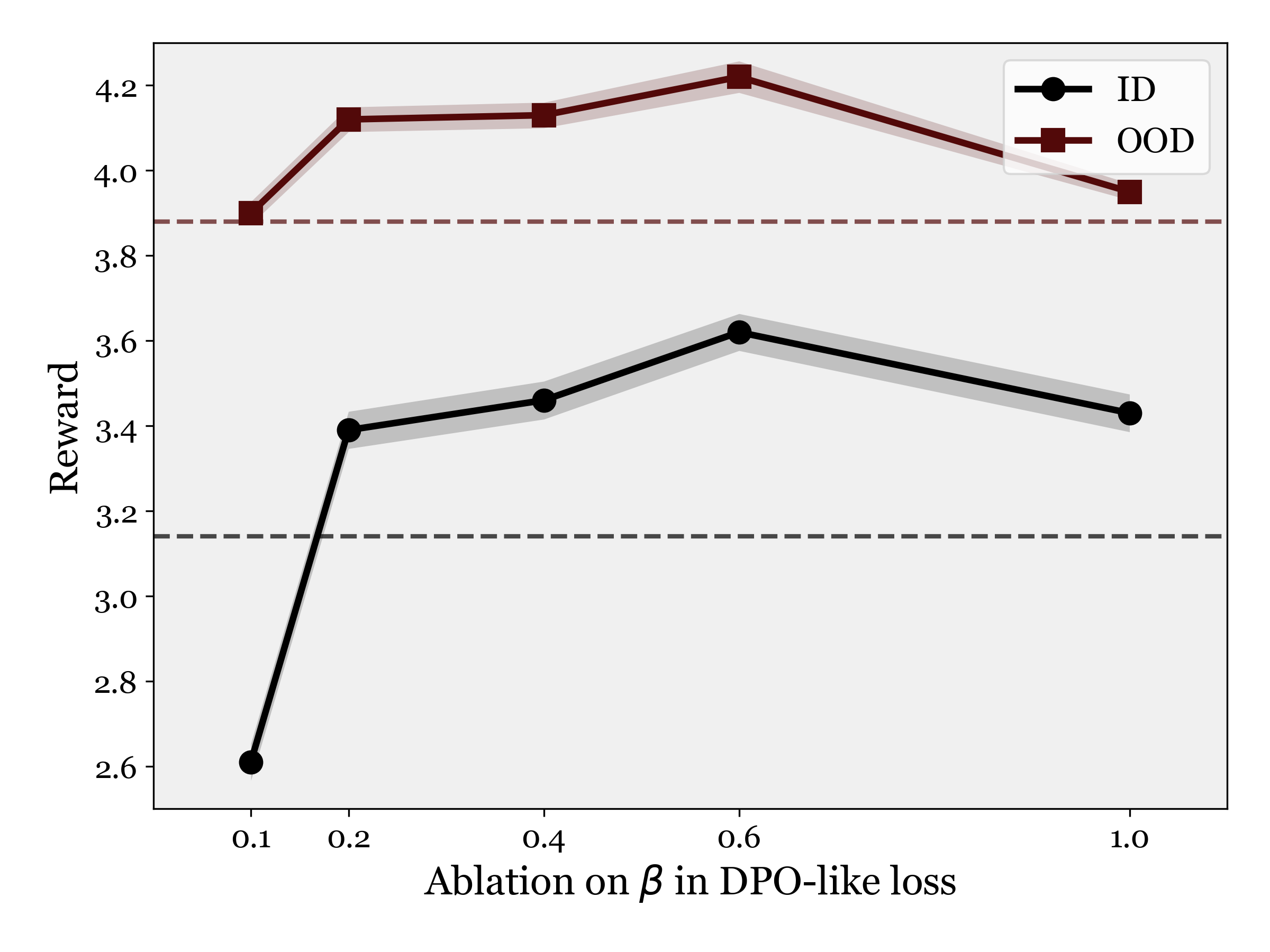}
    \caption{
        Performance comparison of GROOT in the tree-chopping task under different $\beta$ values. The evaluation is conducted in an in-distribution environment.
    }
    \label{fig:beta_selection}
    \vspace{-0.1 in}
\end{figure*}

The experiments in this paper primarily adopt DPO as an example of preference learning within the \alg framework. This algorithm involves the hyperparameter $\beta$. Figure~\ref{fig:beta_selection} illustrates the rationale behind our choice of $\beta$. Although we selected $\beta$ based on a single task \texttt{collect\_wood}(\includegraphics[scale=0.035,valign=c]{sections/assets/minecraft/log.png}) from a single model GROOT, we believe that the effectiveness of this framework ensures strong performance, even if the chosen hyperparameter is not optimal for other tasks or models. The results in Table~\ref{tab:skillforge} further confirm that $\beta=0.6$ is indeed effective.

\section{Subjective Tasks}

\label{climb}

We also conduct experiments based on human subjective preference. The aforementioned experiments are all based on objective environmental feedback in the form of rewards, which can be seen as the ``hard version'' of human preferences. Additionally, we conducted validation experiments under the ``soft version'' as well. Considering the relatively high cost, we only conduct the experiment on $\texttt{explore\_climb}$, which requires the agent to jump up the nearby hill terrain step by step, using human rankings and evaluating through TrueSkill scoring~\citep{trueskill}. The training set contains 40 positive and 40 negative samples, rather than 150 samples as used in objective tasks, due to the human labor cost involved in data annotation. The results are shown in \ref{tab:climb}. Although the confidence in the ranking of the three models' performance is not high, considering the scale of the training data, PGT can still be regarded as demonstrating a certain level of capability in subjective tasks.

\begin{table}[]
\centering
\caption{Performance comparison on the subjective task 
\texttt{explore\_climb}(\includegraphics[scale=0.05,valign=c]{sections/assets/minecraft/Steve.png}) using TrueSkill scores. We benchmark pretrained GROOT, DPO-finetuned GROOT, and GROOT with \alg (GROOT+ in Table~\ref{tab:skillforge}) using 40 trajectory pairs for training and testing.}
\label{tab:climb}
\resizebox{0.9\linewidth}{!}{
\renewcommand\arraystretch{1.1}
\begin{tabular}{@{}cccclccc@{}}
\toprule
\multirow{2}{*}{\includegraphics[scale=0.08,valign=c]{sections/assets/minecraft/Steve.png}} & \multicolumn{3}{c}{In Distribution (ID)} &  & \multicolumn{3}{c}{Out of Distribution (OOD)} \\ \cmidrule(lr){2-4} \cmidrule(l){6-8} 
 & Pretrained & Full & \alg(Ours) &  & Pretrained & Full & \alg(Ours) \\ \midrule
mean($\mu$)$\pm$std($\sigma$) & 24.8$\pm$0.89 & 25.4$\pm$0.91 & \textbf{26.0$\pm$0.91} &  & 25.1$\pm$1.19 & 25.0$\pm$1.18 & \textbf{26.0$\pm$1.23} \\
\bottomrule
\end{tabular}}
\end{table}

\section{Other Preference Learning Algorithms}
\label{sec:ipo}

Our \alg framework consists of data filtering and preference learning. The aforementioned experiments are all based on DPO for convenience, but other preference learning algorithms like SLiC-HF~\citep{slic, slichf} IPO \citep{ipo} are also possible. We experiment them on the latent goal on several tasks and the results are listed in Table~\ref{tab:ipo}. It can be observed that all of them improve task performance across different environments. Different tasks are suited to different algorithms (which may also be related to hyperparameters), but performance almost consistently improves after \alg, and a latent goal with just 512-dimensional parameters is sufficient.

\begin{table}[]
\centering
\caption{\alg with other preference learning algorithms - IPO and SLiC-HF, on GROOT agent.}
\label{tab:ipo}
\resizebox{0.7\linewidth}{!}{
\renewcommand\arraystretch{1.1}
\begin{tabular}{@{}cccccccccc@{}}
\toprule
\multirow{2}{*}{Task} & \multicolumn{4}{c}{In Distribution(ID)} &  & \multicolumn{4}{c}{Out of Distribution(OOD)} \\ \cmidrule(lr){2-5} \cmidrule(l){7-10} 
 & Pretrained & DPO & IPO & SLiC-HF &  & Pretrained & DPO & IPO & SLiC-HF \\ \midrule
\includegraphics[scale=0.043,valign=c]{sections/assets/minecraft/log.png} & 3.14 & \textbf{3.62} & 3.37 & 3.24 &  & 3.88 & \textbf{4.22} & 3.99 & 4.00 \\
\includegraphics[scale=0.045,valign=c]{sections/assets/minecraft/stonecutter.png} & 31.0 & \textbf{44.6} & 42.0 & 37.0 & & 20.0 & 23.4 & 23.0 & \textbf{25.6} \\
\includegraphics[scale=0.06,valign=c]{sections/assets/minecraft/stone.jpg} & 4.91 & \textbf{6.58} & 5.44 & 6.34 & & 3.90 & \textbf{5.38} & 4.70 & 5.29 \\
\includegraphics[scale=0.05,valign=c]{sections/assets/minecraft/hunt.jpg} & 31.2 & 39.8 & 40.6 & \textbf{43.0} & & 20.8 & 21.6 & \textbf{32.8} & 24.4 \\
\includegraphics[scale=0.043,valign=c]{sections/assets/minecraft/pumpkin.png} & 48.3 & 57.8 & \textbf{62.2} & 60.6 & & 16.6 & 25.8 & \textbf{30.6} & 27.6 \\
\includegraphics[scale=0.7,valign=c]{sections/assets/minecraft/obsidian.png} & 42.0 & \textbf{57.2} & 50.4 & 34.4 & & 4.2 & \textbf{8.2} & 4.8 & 3.4 \\ \bottomrule
\end{tabular}}
\end{table}

\end{document}